\documentclass{tlp}

\usepackage{times}
\usepackage[english]{babel}
\usepackage[dvips]{graphicx}
\usepackage{epsfig}
\usepackage{aopmath}

\newcommand{\st}{\medskip\noindent}

\def\ni{\noindent}
\newcommand{\sni}{\smallskip\noindent}

\newtheorem{definition}{Definition}

\def\beq{\begin{equation}}
\def\eeq#1{\label{#1}\end{equation}}
\long\def\COMMENT#1\ENDCOMMENT{\message{(Commented text...)}\par}

\newcommand{\G}{\Gamma}

\def\and{ \ \wedge}

\def\implies{ \supset}

\def\ar{\leftarrow}

\newcommand{\then}{\Rightarrow}

\newcommand{\no}{\hbox{\it not}\ }

\newcommand{\li}{\mbox{\textbf{---\ \,}}}
\newcommand{\lili}{\mbox{\textbf{-----\ \,}}}

\newcommand{\cali}{\mathcal{I}}

\newcommand{\cals}{\mathcal {S}}

\newcommand{\calm}{\mathcal {M}}

\newcommand{\be}{\begin{em}}
\newcommand{\ee}{\end{em}}
\newcommand{\rif}{~\ref}
\newcommand{\as}{``}

\submitted{11 June 2000} \revised{6 November 2003, 15 July 2004}
\accepted{20 December 2004}

\begin{document}
\begin{sloppypar}
\bibliographystyle{acmtrans}

  \title[On the existence of stable models]
        {On the existence of stable models of non--stratified logic programs}

  \author[S. Costantini]
     {Stefania COSTANTINI\\
     Dip. di Informatica, Universit\`a di L'Aquila\\
         via Vetoio Loc. Coppito, L'Aquila, I-67010 Italy\\
         \email{stefcost@di.univaq.it}
         }

\pagerange{\pageref{firstpage}--\pageref{lastpage}}
\volume{\textbf{X} (Y):} \jdate{today} \setcounter{page}{1}
\pubyear{2005}

\label{firstpage}

\maketitle

\begin{abstract}

In this paper we analyze the relationship between cyclic definitions
and consistency in Gelfond-Lifschitz's answer sets semantics
(originally defined as `stable model semantics'). This paper
introduces a fundamental result, which is relevant for Answer Set
programming, and planning. For the first time since the definition
of the stable model semantics, the class of logic programs for which
a stable model exists is given a syntactic characterization. This
condition may have a practical importance both for defining new
algorithms for checking consistency and computing answer sets, and
for improving the existing systems. The approach of this paper is to
introduce a new {\em canonical form} (to which any logic program can
be reduced to), to focus the attention on cyclic dependencies. The
technical result is then given in terms of programs in canonical
form (canonical programs), without loss of generality: the stable
models of any general logic program coincide (up to the language) to
those of the corresponding canonical program. The result is based on
identifying the cycles contained in the program, showing that stable
models of the overall program are composed of stable models of
suitable sub-programs, corresponding to the cycles, and on defining
the \be Cycle Graph. \ee Each vertex of this graph corresponds to
one cycle, and each edge corresponds to one \be handle, \ee which is
a literal containing an atom that, occurring in both cycles,
actually determines a connection between them. In fact, the truth
value of the handle in the cycle where it appears as the head of a
rule, influences the truth value of the atoms of the cycle(s) where
it occurs in the body. We can therefore introduce the concept of a
\be handle path, \ee connecting different cycles. Cycles can be
even, if they consist of an even number of rules, or vice versa they
can be odd. Problems for consistency, as it is well-known, originate
in the odd cycles. If for every odd cycle we can find a handle path
with certain properties, then the existence of stable model is
guaranteed. We will show that based on this results new classes of
consistent programs can be defined, and that cycles and cycle graphs
can be generalized to components and \be component graphs. \ee
\end{abstract}
\newpage

\section{Introduction}

In this paper we analyze the relationship between cyclic definitions
and consistency in Gelfond-Lifschitz's answer sets semantics. As it
is well-known, under the answer set semantics a theory may have no
{\em answer sets}, since the corresponding general logic program may
have no {\em stable models} \cite{GelLif88} \cite{GelLif91}.

We introduce a fundamental result, which is relevant for Answer Set
Programming \cite{MarTru99}, \cite{Nie99} planning \cite{Lif99} and
diagnosis \cite{BaGel03}. For the first time, the class of logic
programs for which a stable model exists is given a syntactic
characterization (the result extends naturally to answer sets
semantics) by providing a necessary and sufficient condition.

While checking for the existence of stable models is as hard a
computational problem (in fact, NP-complete) as planning under
certain assumptions (see \cite{Lib99}), consistency checking is a
good conceptual tool when derivations are based on consistency
arguments. This is the case for instance for all formalizations that
treat goals as constraints over models of the program. Then, being
able to check for the existence of stable models syntactically for
every answer set program can be of help for the logic programming
encodings of planning and diagnosis (like, e.g., those of
\cite{Erd99}, \cite{FabLeoPfe99}, \cite{BBLMP00}, \cite{DNK97} and
\cite{BaGel03}), and for Answer Set Programming in general.

The approach of this paper is to introduce a new {\em canonical
form} to which any logic program can be reduced. The technical
result is then given in terms of programs in canonical form
(canonical programs), without loss of generality. Canonical programs
focus the attention on cyclic dependencies. Rules are kept short, so
as to make the syntactic analysis of the program easier. The stable
models of any general logic program coincide (up to the language) to
those of the corresponding canonical program.

A detailed analysis of the steps involved in reducing programs to
their canonical form has been performed in \cite{CosPro04} and, as
intuition suggests, this transformation is tractable. Nevertheless,
all definitions and results presented in this paper might be
rephrased for general programs without conceptual problems, just at
the expense of a lot of additional details. This means that
reduction to canonical form is not strictly required neither for the
theory, nor for an implementation.

The main result of this paper is a necessary and sufficient
syntactic condition for the existence of stable models. On the one
hand, this condition is of theoretical interest, as it is the first
one ever defined since the introduction of the stable model
semantics in \cite{GelLif88}. On the other hand, it may have a
practical importance both for defining new algorithms for checking
consistency and computing answer sets, and for improving the
existing systems \cite{solvers}.

The result is based on identifying the cycles contained in the
program, on showing that stable models of the overall program are
composed of stable models of suitable sub-programs, corresponding to
the cycles, and on representing the program by means of its \be
Cycle Graph. \ee Each vertex of this graph corresponds to one cycle,
and each edge corresponds to one \be handle, \ee which is a literal
containing an atom that, occurring in both cycles, actually
determines a connection between them. In fact, the truth value of
the handle in the cycle where it appears as the head of a rule
influences the truth value of the atoms of the cycle(s) where it
occurs in the body. We can therefore introduce the concept of a \be
handle path, \ee connecting different cycles. Cycles can be even, if
they consist of an even number of rules, or vice versa they can be
odd. Problems for consistency, as it is well-known, originate in the
odd cycles. If and only if for every odd cycle we can find a
subgraph with certain properties, then the existence of stable
models is guaranteed.

The necessary and sufficient condition that we introduce is
syntactic in the sense that it does not refer either to models or
derivations. Checking this condition requires neither finding the
stable models nor applying the rules of the program. It just
requires to look at the program (represented by the Cycle Graph) and
at the rules composing the cycles. The condition can however be
exploited, so as to obtain: (i) new algorithms for finding the
stable models, which are at least of theoretical interest; (ii) a
new method for consistency checking divided into two steps: a first
step related to the coarse structure of the program, that can be
easily checked on the Cycle Graph so as to rule out a lot of
inconsistent programs, thus leaving only the potentially consistent
ones to be verified in a second step, that can be performed
according to the approach presented here, or in any other way.

We will argue that the approach can also be useful for defining
classes of programs that are consistent by construction, and as a
first step toward a component-based methodology for the construction
and analysis of answer set programs. This by means of a further
generalization of Cycle Graphs to \be Component Graphs, \ee where
vertices are \be components\, \ee consisting of bunches of cycles,
and edges connect different components. We will argue that, in this
framework, components can even be understood as independent agents.

It is useful to notice that in Answer Set Programming graph
representations have been widely used for studying and
characterizing properties of answer set programs, first of all
consistency, and for computing the answer sets. Among the most
important approaches we may mention the Rule Graph \cite{DimTor96}
and its extensions \cite{Lin01} \cite{LinkeSuit} \cite{LinkeRulegr},
and the Extended Dependency Graph \cite{BCDP99}, that we have
considered and compared \cite{Cos01}, \cite{CosDanPro02}. Enhanced
classes of graphs have been recently introduced in order to cope
with extensions to the basic paradigm such as for instance
preferences \cite{LinkePrRep} or nested logic programs
\cite{LinkeNesProg}. However, the Cycle Graph proposed in this paper
is different from all the above-mentioned approaches, since its
vertices are not atoms or rules, but significant subprograms (namely
cycles), and the edges are connections between these subprograms.

\section{Preliminary Definitions}

We consider the standard definition of a (propositional) general
logic program and of well-founded \cite{wfm} and stable model
\cite{GelLif88} and answer set semantics \cite{GelLif91}. Whenever
we mention consistency (or stability) conditions, we refer to the
conditions introduced in \cite{GelLif88}. This section summarizes
some basic definitions, and is intended for readers who are
unfamiliar with the above-mentioned topics.

Assume a language of constants and predicate constants. Assume also
that terms and atoms are built as in the corresponding first-order
language. Unlike classical logic and standard logic programming, no
function symbols are allowed. A rule is an expression of the form:

\beq \rho \: : \: \lambda_0 \ar \lambda_1,\dots,\lambda_m, \no
\lambda_{m+1},\dots,\no \lambda_n \eeq{rule}

\ni where $\lambda_0, \dots \lambda_n$ are atoms and $\no$ is a
logical connective called {\em negation as failure}. The
$\lambda_i$'s are called {\em positive literals}, and the $\no
\lambda_j$'s {\em negative literals} For every rule let us define
$head(\rho) = \lambda_0$ (also called the {\em conclusion} of the
rule), $pos(\rho) = \lambda_1,\dots,\lambda_m$, $neg(\rho) =
\lambda_{m+1},\dots, \lambda_n$ and $body(\rho) = pos(\rho) \: \cup
\: neg(\rho)$  (also called the {\em conditions} of the rule). If
$body(\rho) = \emptyset$ we refer to $\rho$ as a {\em unit rule }
(w.r.t. non-unit rules), or a {\em fact}. We will say that
$head(\rho)$ {\em depends on}, or {\em is defined in terms of}, the
literals in $body(\rho)$.

A general logic program $\Pi$ (or simply \as logic program'') is
defined as a collection of rules. In the rest of this paper, we rely
on the assumption that the order of literals in the body of rules is
irrelevant. Rules with variables are taken as shorthand for the sets
of all their ground instantiations and the set of all ground atoms
in the language of a program $\Pi$ will be denoted by ${\rm I
\mkern-4mu B}_{\Pi}$.

\subsection{Semantics}
For the sake of simplicity, we give here the definition of {\em
stable model} instead of that of answer set, which is an extension
given for programs that contain the explicit negation operator
$\neg$. In fact, this is not going to make a difference in the
context of this work, and we will often interchange the terms \as
stable models'' and \as answer sets''. Intuitively, a stable model
is a possible view of the world that is {\em compatible} with the
rules of the program. Rules are therefore seen as constraints on
these views of the world.

\ni Let us start by defining stable models of the subclass of
positive programs, i.e. those where, for every rule $\rho$,
$neg(\rho) = \emptyset$.

\begin{definition} (Stable Models of positive logic programs)\newline \label{posStableModel}

\ni The {\em stable model} $a(\Pi)$ of a positive program $\Pi$ is
the smallest subset of ${\rm I \mkern-4mu B}_{\Pi}$ such that for
any rule (\ref{rule}) in $\Pi$:

\beq \lambda_1,\ldots,\lambda_m \in a(\Pi) \then \lambda_0 \in
a(\Pi) \eeq{posStable}

\end{definition}

\ni Positive programs have a unique stable model, which coincides
with its minimal model, that can also be obtained applying other
semantics; then, positive programs are unambiguous. The stable model
of a positive program can be obtained as the fixed point of the {\it
immediate consequence operator}
\[T_{\Pi}(I)=\{\lambda: \exists \rho\in\Pi\ \mbox{\ s.t.\ } \ \lambda
= head(\rho) \wedge \ pos(\rho)\subseteq I\}\] The iterated
application of $T_{\Pi}$ from $\emptyset$ on (i.e.,
$T_{\Pi}(\emptyset),T^2_{\Pi}(\emptyset),\dots$) is guaranteed to
have a fixed point, which corresponds to the unique stable model
(answer set) of $\Pi$.

A set of atoms $\cals$ is a stable model of an (arbitrary) program
if it is a minimal model and every atom $\alpha \in \cals$ is
supported by some rule of the program. With respect of negation, if
we assume $\cals$ to be a stable model: (i) no atom can belong to
$\cals$, which is derived by means of a rule with a condition $\no
\alpha$ where $\alpha$ is true in $\cals$, i.e. $\alpha \in \cals$;
(ii) all literals $\no \beta$ in the body of rules where $\beta$ is
false in $\cals$ are, of course, true in $\cals$. Consequently, in
order to check whether $\cals$ actually is a stable model, all
negations can be deleted according to the these criteria, in order
to apply the above formulation for positive programs.

\begin{definition} (Stable Model of arbitrary logic programs)\newline \label{StableModel}

\ni Let $\Pi$ be a logic program. For any set $\cals$ of atoms, let
$\Pi^{\cals}$ be a program obtained from $\Pi$ by deleting

\begin{description}
\item[\rm{(i)}] each rule that has a formula ``$\no\ \lambda$'' in its body with
$\lambda \in S$;

\st

\item[\rm{(ii)}] all formulae of the form ``$\no\ \lambda$'' in the bodies
of the remaining rules.

\end{description}

\ni Since $\Pi^{\cals}$ does not contain $\no$, its stable model is
already defined. If this stable model coincides with $\cals$, then
we say that $\cals$ is a {\em stable model} of $\Pi$. Precisely, a
stable model of $\Pi$ is characterized by the equation:

\beq \cals \ =\ a(\Pi^{\cals}). \eeq{stableModel}

\end{definition}

\ni The $\G$ operator, introduced by Gelfond and Lifschitz in
\cite{GelLif88}, is defined as $\G(\Pi,{\cals})\ =\ a(\Pi^{\cals})$.
When $\Pi$ is fixed, we may drop the first parameter and refer to
$\G$ as a function from the powerset of ${\rm I \mkern-4mu B}_{\Pi}$
to itself. In practice however, stable models are not computed by
applying $\G$ to all subsets of ${\rm I \mkern-4mu B}_{\Pi}$. Answer
set solvers \cite{solvers} in fact apply more effective and smarter
algorithms.

Stable models are minimal models of $\Pi$ in the classical sense,
but the converse does not hold. Then, a program may have no stable
models. In general a program has several stable models, and programs
with a unique stable model are called {\em categorical}. In this
paper, consistency means existence of stable models (or,
equivalently, of answer sets). Conventionally, \as an atom being
true'' means \as an atom being in a stable model''. Whenever we
consider a set of atoms $\cali$, we implicitly mean $\cali \subseteq
{{\rm I \mkern-4mu B}_{\Pi}}$. We say that a literal $\alpha$
(respectively $\no \alpha$) is true w.r.t. $\cali$ if $\alpha \in
\cali$ (respectively $\cali$ if $\alpha \not\in \cali$).

\subsection{The well-founded semantics}\label{well-founded}
The well-founded semantics of \cite{wfm} assigns to a logic program
$\Pi$ a unique, three-valued model, called well-founded model and
denoted by $WFS(\Pi)=\langle W^{+},W^{-}\rangle$, where $W^{+}$ and
$W^{-}$ are disjoint. Intuitively, $W^{+}$ is the set of atoms
deemed true, $W^{-}$ is the set of atoms deemed false, while atoms
belonging to neither set are deemed {\em undefined.}

The reduction of a program to its {\em canonical form} that we
discuss later is based on a preliminary simplification of the
program w.r.t. the well-founded semantics. The result of this is a
compact version (or {\em reduct}) of the program which is {\em
WF-irreducible}, where

\begin{definition}
A program $\Pi$ is {\em WF-irreducible} if  and only if
$WFS(\Pi)=\langle\emptyset,\emptyset\rangle$.
\end{definition}

For general logic programs, atoms with truth value ``undefined''
under the well-founded semantics are exactly the atoms which are of
interest for finding the stable models. This is a consequence of the
fact that all stable models of a program \textit{extend} its
well-founded model, i.e., every literal which is true (resp. false)
in the well-founded model is also true (resp. false) in all stable
models. The stable models of the original program can be easily
obtained from the stable models of the WF-irreducible reduct
\cite{Cos95}, and vice versa, if the reduct has no stable models the
same holds for the original program.

For instance, for program

\st $
\begin{array}{l}
p \ar \no p, \no q
\end{array} $

with well-founded model $\langle \emptyset; \{q\}\rangle$ where atom
$p$ has truth value ``undefined'', we get the simplified
WF-irreducible reduct $p \ar \no p$ by getting rid of a literal
which is true under the well-founded semantics, and thus is true in
all stable models (if any exists). The reduct has no stable models,
like the original program.

\section{Cycles and Handles}
\label{CyclesandHandles}

The results on consistency checking that we will present later are
based on identifying the negative cycles contained in the program,
on showing that stable models of the overall program are composed of
stable models of suitable sub-programs, corresponding to the
negative cycles, and on representing the program by means of its
{\em Cycle Graph}. In this section we define when a set of rules
constitutes a negative cycle (or simply \as cycle''), which kinds of
cycles we may have and how cycles can be understood to be connected
to each other.

\begin{definition}
A set of rules $C$ is called a {\em negative cycle}, or for short a
{\em cycle}, if it has the following form:

\st $
\begin{array}{l}
\lambda_1 \ar \no \lambda_2, \Delta_1\\
\lambda_2 \ar \no \lambda_3, \Delta_2\\
\dots\\
\lambda_n \ar \no \lambda_1, \Delta_n
\end{array} $

\st where $\lambda_1,\ldots ,\lambda_n$ are distinct atoms. Each
$\Delta_i$, $i \leq n$, is a (possibly empty) conjunction
$\delta_{i_1}, \ldots, \delta_{i_h}$ of literals (either positive or
negative), where for each $\delta_{i_j}$, $i_j \leq i_h$,
$\delta_{i_j} \neq \lambda_i$ and $\delta_{i_j} \neq \no \lambda_i$.
The $\Delta_i$'s are called the {\em AND handles} of the cycle. We
say that $\Delta_i$ is an AND handle for atom $\lambda_i$, or,
equivalently, an AND handle referring to $\lambda_i$.
\end{definition}

We say that $C$ has size $n$ and it is even (respectively odd) if
$n=2k$, $k \geq 1$ (respectively, $n=2k+1$, $k \geq 0$). For $n=1$
we have the (odd) self-loop $\lambda_1 \ar \no\lambda_1,\Delta_1$.
In what follows, $\lambda_{i+1}$ will denote $\lambda_{(i+1) mod
n}$, i.e., $\lambda_{n+1} = \lambda_{1}$.

A {\em positive cycle} is similar to a negative cycle, except that
we have positive literals $\lambda_i$'s in the body of rules instead
of negative ones. In the rest of the paper we will consider negative
cycles unless differently specified explicitly.

For any cycle $C$, we will denote by $Composing\_atoms(C)$ the set
$\{\lambda_1,\ldots ,\lambda_n\}$, i.e., the set of atoms \be
involved \ee in cycle $C$. We say that the rules listed above are
\be involved \ee in the cycle, or \be form \ee the cycle. In the
rest of the paper, sometimes it will be useful to see
$Composing\_atoms(C)$ as divided into two subsets, that we indicate
as two {\em kinds} of atoms: the set of the {\em Even\_atoms(C)}
composed of the $\lambda_i$'s with $i$ even, and the set {\em
Odd\_atoms(C)}, composed of the $\lambda_i$'s with $i$ odd.

Conventionally, in the rest of the paper $C$ and $C_i$  denote
cycles in general, $OC$ and $OC_i$ denote odd cycles, and $EC$ or
$EC_i$ denote even cycles.

In the following program for instance, there is an odd cycle that we
may call $OC_1$, with composing atoms $\{e, f, g\}$ and an even
cycle that we may call $EC_1$, with composing atoms $\{a,b\}$.

\st $
\begin{array}{l}
\li OC_1\\
e \ar \no f, \no a\\
f \ar \no g, b\\
g \ar \no e\\
\li EC_1\\
a \ar \no b\\
b \ar \no a
\end{array} $

\ni $OC_1$ has an AND handle $\no a$ referring to $e$, and an AND
handle $b$ referring to $f$.

Notice that the sets of atoms composing different cycles are not
required to be disjoint. In fact, the same atom may be involved in
more than one cycle. The set of atoms composing a cycle can even be
a proper subset of the set of atoms composing another cycle, like in
the following program, where there is an even cycle $EC_1$ with
composing atoms $\{a,b\}$, since $a$ depends on $\no b$ and $b$
depends on $\no a$, but also an odd cycle $OC_1$ with composing atom
$\{a\}$, since $a$ depends on $\no a$.

\st $
\begin{array}{l}
\li EC_1\\
\lili OC_1\\
a \ar \no a, \no b\\
b \ar \no a
\end{array} $

Here, $OC_1$ has an AND handle $\no b$ referring to $a$,
but, symmetrically, $EC_1$ has an AND handle $\no a$ referring to $b$.

Thus, it may be the case that a handle of a cycle $C$ contains an atom
$\alpha$ which is involved
in $C$ itself, because $\alpha$ is also involved in some other cycle
$C_1$.

\begin{definition}\label{or-one-handle}
A rule is called an \be auxiliary rule of cycle \ee $C$
(or, equivalently, \be to \ee cycle $C$)
if it is of this form:

\noindent
$\lambda_i \ar \Delta$

\noindent where $\lambda_i \in$ Composing\_Atoms($C$), and $\Delta$
is a non-empty conjunction $\delta_{i_1}, \ldots, \delta_{i_h}$ of
literals (either positive or negative), where for each
$\delta_{i_j}$, $i_j \leq i_h$, $\delta_{i_j} \neq \lambda_i$ and
$\delta_{i_j} \neq \no \lambda_i$. $\Delta$ is called an {\em OR
handle} of cycle $C$ (more specifically, an OR handle for
$\lambda_i$ or, equivalently, and OR handle referring to
$\lambda_i$).

\end{definition}

A cycle may possibly have several auxiliary rules, corresponding
to different OR handles. In the rest of this paper, we will call
\be Aux(C) \ee the set of the auxiliary rules of a cycle $C$.

In the following program for instance, there is an odd cycle
$OC_1$ with composing atoms $\{c,d,e\}$ and an even cycle $EC_1$
with composing atoms $\{a,b\}$. The odd cycle has three auxiliary
rules.

\st $
\begin{array}{l}
\li OC_1\\
c \ar \no d\\
d \ar \no e\\
e \ar \no c\\
\lili Aux(OC_1)\\
c \ar \no a\\
d \ar \no a\\
d \ar \no b\\
\li EC_1\\
a \ar \no b\\
b \ar \no a
\end{array} $

In particular, we have $Aux(OC_1) = \{c \ar \no a, d \ar \no a, d
\ar \no b\}$.

In summary, a cycle may have some AND handles, occurring in one or
more of the rules that form the cycle itself, and also some OR
handles, occurring in its auxiliary rules. Cycles and handles can be
unambiguously identified on the Extended Dependency Graph (EDG) of
the program \cite{BCDP99}.

A cycle may also have no AND handles and no OR handles, i.e., no
handles at all, in which case it is called \be unconstrained.\ee
The following program is composed of unconstrained cycles (in
particular, there is an even cycle involving atoms $a$ and $b$,
and an odd cycle involving atom $p$).

\st $
\begin{array}{l}
\li EC_1\\
a \ar \no b\\
b \ar \no a\\
\li OC_1\\
p \ar \no p
\end{array} $

Notice that the basic definition of a cycle corresponds to that of a
{\em negative cycle} in the {\em atom Dependency Graph} as defined
in \cite{Fag94}. However, as discussed in \cite{Cos01}, on the
dependency graph it is impossible to identify the handles, and there
are different programs with different answer set, but the same
dependency graph. The handles can be identified unambiguously on the
{\em Extended Dependency Graph} as defined and discussed in
\cite{BCDP99} and \cite{CosDanPro02}.


\section{Canonical programs}
\label{canonicalPrograms}

In order to analyze the relationship between cycles, handles and
consistency, below we introduce a \be canonical form \ee for logic
programs. This canonical form is new, and is introduced with the
general objective of simplifying the study of formal properties of
logic programs under the Answer Set semantics. Rules in \be
canonical programs \ee are in a standard format, so as to make
definitions and proofs cleaner and easier to read. There is however
no loss of generality, since, as proved in the companion paper
\cite{CosPro04}, any logic program can be {\em reduced} to a
canonical program, and that stable models of the original program
can be easily obtained from the stable models of its canonical
counterpart.

\begin{definition}
\label{canonicalProg}
A logic program $\Pi$ is in canonical form
(or, equivalently, $\Pi$ is a canonical program) if it is
$WF$-irreducible, and fulfills the following syntactic conditions.

\begin{enumerate}
\item $\Pi$ does not contain positive cycles;
\item every atom in $\Pi$ occurs both in the head of some rule and in the body of some other (possibly the same) rule;
\item every atom in $\Pi$ is involved in some cycle;
\item each rule of $\Pi$ is either involved in a cycle,
or is an auxiliary rule of some cycle;
\item each handle of a cycle $C$ consists of exactly one literal, either $\alpha$ or $\no \alpha$, where
atom $\alpha$ does not occur in $C$.
\end{enumerate}
\end{definition}

Since the above definition requires handles to consist of just one
literal, it implies that in a canonical program $\Pi$ : (i) the
body of each rule which is involved in a cycle consists of either
one or two literals; (ii) the body of each rule which is an
auxiliary rule of some cycle consists of exactly one literal.

Nothing prevents a rule to be at the same time involved in a cycle,
and auxiliary to some other cycle. In this case however, the
definition requires the rule to have exactly one literal in the
body, i.e., the rule cannot have an AND handle.

All definitions and results introduced in the rest of the paper
might be rephrased for the general case, but the choice of referring
to canonical programs is a significant conceptual simplification
that leads without loss of generality to a more readable and
intuitive formalization. Notice for instance that in canonical
programs the problem of consistency arises only in cycles containing
an odd number of rules, since cycles do not have non-negated
composing atoms.

Although for a detailed discussion we refer to \cite{CosPro04}, it
is important to recall that canonical programs are $WF$-irreducible.
For instance, the program

\st $
\begin{array}{l}
p \ar \no p, q\\
q \ar \no q, p
\end{array} $

may look canonical, while it is not, since it has a non-empty
well-founded model $\langle \emptyset ; \{p,q\} \rangle$. In
particular, since there are no undefined atoms, the set of true
atoms of the well-founded model (in this case $\emptyset$) coincides
(as it is well-known) with the unique stable model.

Similarly, the program

\st $
\begin{array}{l}
q \ar \no q\\
q \ar p
\end{array} $

may look canonical, while it is not, since it has a non-empty
well-founded model $\langle \emptyset ; \{p\} \rangle$. Atom $q$ is
undefined. The corresponding canonical program is $q \ar \no q$
that, like the original program, has no stable models. The second
rule is dropped by canonization, since its condition is false w.r.t.
the well-founded model. The program

\st $
\begin{array}{l}
q \ar p\\
p \ar \no r\\
r \ar \no q
\end{array} $

is not canonical because atom $p$ is not involved in any cycle. In
fact, in order to be involved in a cycle an atom must occur in the
head of some rule but, also, its negation must occur in the body of
some, possibly different, rule. Here, atom $p$ forms an
(inessential) intermediate step between the two atoms $q$ and $r$
that actually form a cycle. The canonical form of the program is $q
\ar \no r, r \ar \no q$. Given the stable models $\{q\}$ and $\{r\}$
of the canonical program, the stable models $\{p,q\}$ and $\{r\}$ of
the original program can be easily obtained, since the truth value
of $p$ directly depends on that of $r$.

In the following, let the program at hand be a logic program $\Pi$
in canonical form, unless differently specified explicitly. Let
$C_1,\ldots,C_w$ be all the cycles occurring in $\Pi$ (called \be
the composing cycles \ee of $\Pi$). Whenever we will refer to $C$,
$C_1$, $C_2$, $C_i$ etc. we implicitly assume that these are cycles
occurring in $\Pi$.

\section{Active handles}

In this section we make some preliminary steps toward establishing a
connection between syntax (cycles and handles) and semantics
(consistency of the program). Truth or falsity of the atoms
occurring in the handles of a cycle (w.r.t. a given set of atoms)
affects truth/falsity of the atoms involved in the cycle. As we
discuss at length in the rest of the paper, this creates the
conditions for stable models to exist or not, and these conditions
can be checked on the Cycle Graph of the program. This idea is
formalized in the following definitions of \be active handles, \ee
that will be frequently used in the rest of the paper.

\begin{definition}
Let $\cali$ be a set of atoms. An AND handle $\Delta$ of cycle $C$
is {\em active} w.r.t. $\cali$ if it is false w.r.t. $\cali$. We say
that the rule where the handle occurs has an {\em active AND
handle}. An OR handle $\Delta$ of cycle $C$ is {\em active} w.r.t.
$\cali$ if it is true w.r.t. $\cali$. We say that the rule where the
handle occurs has an {\em active OR handle}.
\end{definition}

Assume that $\cali$ is a model. We can make the following
observations. (i) The head $\lambda$ of a rule $\rho$ with an
active AND handle is not required to be true in $\cali$. (ii) The
head of a rule $\lambda \ar \Delta$ where $\Delta$ is an active
OR handle is necessarily true in $\cali$: since the body is true,
the head $\lambda$ must also be true.

Observing which are the active handles of a cycle $C$ gives relevant
indications about whether a set of atoms $\cali$ is a stable model.

Consider for instance the following program:

\st $
\begin{array}{l}
\li OC_1\\
p \ar \no p, \no a\\
\li EC_1\\
a \ar \no b\\
b \ar \no a\\
\li OC_2\\
q \ar \no q\\
\lili Aux(OC_2)\\
q \ar f\\
\li EC_2\\
e \ar \no f\\
f \ar \no e
\end{array} $

\ni The sets of atoms $\{a,f,q\}$, $\{a,e,q\}$, $\{b,p,f,q\}$ $\{b,
p,e,q\}$ are minimal models. Consider the set of atoms $\{a,f,q\}$:
both the AND handle $\no a$ of cycle $p \ar \no p, \no a$ and the OR
handle $f$ of cycle $q \ar \no q$ are active w.r.t. this set of
atoms. $\{a,f,q\}$ is a stable model, since atom $p$ is forced to be
false, and atom $q$ is forced to be true, thus avoiding the
inconsistencies. In all the other minimal models instead, one of the
handles is not active. I.e., either literal $\no a$ is true, and
thus irrelevant in the context of a rule which is inconsistent, or
literal $f$ is false, thus leaving the inconsistency on $q$. These
minimal models are in fact not stable.

In conclusion, the example suggests that for a minimal model $\calm$
to be stable, each odd cycle must have an active handle. Formally:

\begin{theorem}
\label{prel} Let $\Pi$ be a program, and let $\calm$ be a minimal
model of $\Pi$. $\calm$ is a stable model only if each odd cycle
$OC_i$ occurring in $\Pi$ has an active handle w.r.t. $\calm$.
\end{theorem}
\begin{proof}
Since $\calm$ is stable, for each $A \in \calm$ there exists a rule
in $\Pi$ with head $A$ and body which is true w.r.t. $\calm$, i.e.,
a rule which \be supports \ee $A$. Let $x\ mod\ y$ be (as usual) the
remainder of the integer division of $x$ by $y$.

\ni Assume that $\calm$ is stable, but there is an odd cycle without
active handles, composed of atoms $\lambda_1,\ldots ,\lambda_n$,
where $n$ is odd. Take a $\lambda_i$,  and assume first that
$\lambda_i \in \calm$. Since there is no active OR handle, each
$\lambda_i$ can possibly be supported only by the corresponding rule
in the cycle. By definition of cycle, this rule has the form:
\[\lambda_i \ar \no \lambda_{(i+1)\ mod\ n}, \Delta_i\] Since
there are no active AND handles, then all $\Delta$'s are true w.r.t.
$\calm$. For $\lambda_i$ to be supported, $\no \lambda_{(i+1)\ mod\
n}$ should be true as well, i.e., $\lambda_{(i+1)\ mod\ n}$ should
be false. The rule for $\lambda_{(i+1)\ mod\ n}$ has the form:
\[\lambda_{(i+1)\ mod\ n} \ar \no \lambda_{(i+2)\ mod\ n},
\Delta_{(i+1)\ mod\ n}\] Since $\Delta_{(i+1)\ mod\ n}$ is true
w.r.t. $\calm$, for $\lambda_{(i+1)\ mod\ n}$ to be false, $\no
\lambda_{(i+2)\ mod\ n}$ should be false as well, i.e.,
$\lambda_{(i+2)\ mod\ n}$ should be true. By iterating this
reasoning, $\lambda_{(i+3)\ mod\ n}$ should be false, etc. In
general, $\lambda_{(i+k)\ mod\ n}$ should be false w.r.t. $\calm$
with $k$ odd, and true with $k$ even. Then, since the number $n$ of
the composing atoms is odd, $\lambda_{(i+n)\ mod\ n}$ should be
false w.r.t. $\calm$, but $\lambda_{(i+n)\ mod\ n} = \lambda_i$,
which is a contradiction. Assume now that $\lambda_i \not\in \calm$.
Then, $\no \lambda_i$ is true w.r.t. $\calm$, and thus, since the
corresponding AND handle is not active, $\lambda_{(i-1)\ mod\ n}$ is
supported and should belong to $\calm$. Consequently, we should have
$\lambda_{(i-2)\ mod\ n}\not\in \calm$. In general, $\lambda_{(i-k)\
mod\ n}$ should be true w.r.t. $\calm$ with $k$ odd, and false with
$k$ even. Then, since the number $n$ of the composing atoms is odd,
$\lambda_{(i-n)\ mod\ n}$ should be true w.r.t. $\calm$, but
$\lambda_{(i-n)\ mod\ n} = \lambda_i$, which is again a
contradiction.
\end{proof}

Another thing that the above example shows is that the stable model
$\{a,f,q\}$ of the overall program is actually the union of the
stable model $\{a\}$ of the program fragment $OC_1 \cup EC_1$ and of
the stable model $\{f,q\}$ of the program fragment $OC_2 \cup
Aux(OC_2) \cup EC_2$. This is not by chance, and in the next
sections we will study how to relate the existence of stable models
of the overall program to the existence of stable models of the
composing cycles.

In order to do so, some preliminary definitions about handles are in
order. It is useful to collect the set of handles of a cycle into a
set, where however each handle is annotated so as to keep track of
its \be kind. \ee I.e., we want to remember whether a handle is an
OR handle or an AND handle of the cycle.

\begin{definition}
Given cycle $C$, the set $H_C$ of the handles of $C$ is defined as
follows, where $\beta \in Composing\_Atoms(C)$:

\st $
\begin{array}{lllll}
H_C & = & \{(\Delta : AND : \beta)\, | \, \Delta \mbox{\, is an AND
handle of \,} C \mbox{\, referring to \,} \beta \}
& \cup \\
& & \{(\Delta : OR : \beta)\, | \, \Delta \mbox{\, is an OR handle
of \,} C \mbox{\, referring to \,} \beta \}
\end{array}
$
\end{definition}

Whenever we need not care about $\beta$ we shorten $(\Delta : K :
\beta)$ as $(\Delta : K)$, $K$ = AND/OR. We call ``handles'' the
expressions in both forms, and whenever necessary we implicitly
shift from one form to the other one. Informally, we will say for
instance ``the OR (resp. AND) handle $\Delta$ of $\beta$'' meaning
$(\Delta : OR : \beta)$ (resp. $(\Delta : AND : \beta)$).

In general however the indication of $\beta$ is necessary. In fact,
different atoms of a cycle may have handles with the same $\Delta$,
but although active/not active at the same time, they may affect the
existence of stable models differently. Take for instance the
following program with the indication of the composing cycles:

\st $
\begin{array}{l}
\li OC_1\\
q \ar \no q, e\\
q \ar \no f\\
\li OC_2\\
a \ar \no b, \no e\\
b \ar \no c, \no f\\
c \ar \no a, \no e\\
\li OC_3\\
p \ar \no p, \no e\\
\li EC_1\\
e \ar \no f\\
f \ar \no g
\end{array} $

\ni we have $H_{OC_1} = \{(e:AND:q), (\no f:OR:q)\}$, $H_{OC_2} =
\{(\no e:AND:a), (\no f:AND:b), (\no e:AND:c)\}$, $H_{OC_3} = \{(\no
e:AND:p)\}$, $H_{EC_1} = \emptyset$. Handle $(\no e : AND)$ occurs
several times, even twice in cycle $OC_2$, referring to different
atoms. Notice that the same literal $\Delta$ may occur both in AND
and OR handles. E.g., $\no f$ occurs both in an AND handle (of
$OC_2$) and in and OR handle (of $OC_1$). Notice also that the same
atom $\alpha$ may appear in literals $\alpha$ and $\no \alpha$ that
occur in different handles. E.g., $f$ occurs in an OR handle of
$OC_1$, and $\no f$ occurs in and AND handle of $OC_2$.

Given any subset $Z$ of $H_C$, it is useful to identify the set of
atoms occurring in the handles belonging to $Z$.

\begin{definition}
Let $Z \subseteq H_C$. The set of the atoms occurring in the handles
belonging to $Z$ is defined as follows.

\st $
\begin{array}{lllll}
Atoms(Z) & = & \{ \alpha \, | \, (\alpha : K) \in Z \} \,\,\, \cup \\
& & \{ \alpha \, | \, (\no \alpha : K) \in Z \}
\end{array}
$
\end{definition}

If for instance we take $Z = H_{OC_1}$, we have $Atoms(H_{OC_1}) =
\{e, f\} $.

Given any subset $Z$ of $H_C$, it is useful to state which are the
atoms that are required to be true, in order to make all the handles
in $Z$ active (implicitly, to this aim all the other atoms are
required to be false).

\begin{definition}
Let $Z \subseteq H_C$. The set of atoms $ActivationAt_C(Z) \subseteq
Atoms(Z)$ is defined as follows.

\st $
\begin{array}{lllll}
ActivationAt_C(Z) & = & \{ \alpha \, | \, (\alpha : OR) \in Z \} & \cup \\
& & \{ \alpha \, | \, (\no \alpha : AND) \in Z \}
\end{array}
$
\end{definition}

If for instance we take $Z = H_{OC_3}$, we have
$ActivationAt(H_{OC_1}) = \{e\} $.

Vice versa, any subset $V$ of $Atoms(H_C)$ corresponds to a subset
of the handles of $C$ that become active, if atoms in $V$ are true.

\begin{definition}
Let $V \subseteq Atoms(H_C)$.

\st $
\begin{array}{lllll}
Active_C(V) & = & \{(\Delta : AND)\, | \, \Delta = \no \alpha, \alpha \in V \} \,\,\, \cup \\
& & \{(\Delta : OR)\, | \, \Delta = \alpha, \alpha \in V \}
\end{array}
$
\end{definition}

If for instance we take $V = \{e\}$ for cycle ${OC_3}$, we have
$Active_{OC_3}(\{e\}) = \{(\no e: AND)\} $.

Finally, it is useful to introduce a short notation for the union
of different sets of rules.

\begin{definition}
Let $I_1,\ldots I_q$ be sets of rules. As a special case, some of
the $I_j$'s can be sets of atoms, where each atom $\beta \in I_j$
is understood as a fact
 $\beta \ar$. By $I_1 + \ldots + I_q$ we mean the program consisting
of the union of all the rules belonging to $I_1,\ldots I_q$.
\end{definition}

\section{Cycle, handles and existence of stable models}

In this and the following sections we proceed further toward a
framework that relates cycles, handles and active handles to the
existence of stable models. This relation is far from obvious, as
demonstrated by the following simple program.

\st $
\begin{array}{l}
\li OC_1\\
p \ar \no p, \no a\\
\li EC_1\\
a \ar \no b\\
b \ar \no a\\
\li OC_2\\
q \ar \no q, \no b
\end{array} $

In this case, we have only one even cycle, and we might consider the
program fragments: (i) $OC_1 \cup EC_1$ with stable model $\{a\}$,
based on the active handle $\no a$; (ii) $OC_2 \cup EC_1$ with
stable model $\{b\}$, based on the active handle $\no b$.
Unfortunately, the union $\{a,b\}$ of the stable models of the two
program fragments is a minimal model but it is not stable. In fact,
neither atom $a$ nor atom $b$ is supported. This is because the
unconstrained even cycle $EC_1$, taken per se, has stable models
$\{a\}$ and $\{b\}$, which are alternative and cannot be merged: the
rules of this cycle in fact state that $a$ holds if $b$ does not
hold, and vice versa. Thus, $EC_1$ cannot provide active handles for
both the odd cycles.

Then, if we want to check whether a minimal model is stable, we not
only have to check that every odd cycle has an active handle w.r.t.
that model, but also that these handles do not enforce contradictory
requirements on the even cycles. We can try to build a stable model
of the overall program out of the stable models of the composing
cycles, taking however care of avoiding inconsistencies on the
handles.

Consider a cycle $C_i$ occurring in $\Pi$ together with its
auxiliary rules, i.e., consider the set of rules $C_i + Aux(C_i)$
and take it as an independent program. Notice that this program is
not canonical, since there are atoms that do not appear in the
conclusion of rules: these are exactly the atoms occurring in the
handles of $C_i$, i.e., the atoms in $Atoms(H_{C_i})$. Take a set
$X_i \subseteq Atoms(H_{C_i})$, and assume to add atoms in $X_i$ as
facts to $C_i + Aux(C_i)$.

\begin{definition}
Let $C_i$ be a cycle. Let $X_i \subseteq H_{C_i}$. The general logic
program $C + Aux(C) + X_i$ is called an \be extended cycle \ee of
$C_i$ corresponding to $X_i$.
\end{definition}

Depending on the active handles $Active_{C_i}(X_i)$ corresponding to
$X_i$, the extended cycle $C_i + Aux(C_i) + X_i$ may or may not be
consistent.

\begin{definition}
\label{partialstable} Let $C_i$ be a cycle occurring in $\Pi$. We
say that $C_i$ has \be partial stable models \ee if $\exists X_i
\subseteq Atoms(H_{C_i})$ such that the corresponding extended cycle
$C_i + Aux(C_i) + X_i$ is consistent. Given a stable model $S_{C_i}$
of $C_i + Aux(C_i) + X_i$, the set $X_i$ is called a {\em positive
base} for $S_{C_i}$, while the set $X_i^{-} = Atoms(H_{C_i})
\setminus X_i$ is called a {\em negative base} for $S_{C_i}$. The
couple of sets $\langle X_i, X_i^{-} \rangle$ is called a {\em base}
for $S_{C_i}$. We say that the $S_{C_i}$'s are {\em partial stable
models} for $\Pi$ relative to $C_i$.
\end{definition}

Atoms in $X_i$ are added as facts in order to simulate that we
deduce them true in some other part of the program. Symmetrically,
atoms in $X_i^{-}$ are supposed not to be concluded true anywhere in
the program. The positive base $X_i$ may be empty: in this case, all
the atoms occurring in the handles are supposed to be false.
Clearly, there may be no partial stable models relative to a cycle
$C_i$ or there may be several ones. However, partial stable models
of cycles are related to stable models of the overall program.

\begin{lemma}
\label{restriction} \label{restriction} Let $\Pi$ be a program,
$C_i$ be one of its composing cycle, and $\cali$ be a stable model
of $\Pi$. Let $X_i = \cali \cap Atoms(H_{C_i})$. Then, the
restriction $S_i$ of $\cali$ to the atoms involved in the extended
cycle $P_i = C_i + Aux(C_i) + X_i$ is a partial stable model of
$P_i$.
\end{lemma}
\begin{proof}
Notice that all non-unit rules of $P_i$ are also rules of $\Pi$.
Notice also that for every atom $\alpha$ occurring in $P_i$ as the
head of a non-unit rule, $P_i$ contains all rules of $\Pi$ with head
$\alpha$: as $\Pi$ is canonical, these rules are by
Definition\rif{canonicalProg} either in $C_i$ or in $Aux(C_i)$.
Assume that $S_i$ is not a stable model of $P_i$. This means that,
after applying the reductions specified in
Definition\rif{StableModel}, we obtain a positive program
${P_i}^{S_i}$ where either (i) there exists atom $\alpha \in S_i$
that is not a consequence of ${P_i}^{S_i}$ or (ii) there exists atom
$\beta$ which is a consequence of ${P_i}^{S_i}$, but $\beta \not\in
S_i$. In situation (i), it means that we have canceled all rules
with head $\alpha$, because they contain a negative literal which is
false w.r.t. $S_i$. But, as we have included in $S_i$ all the atoms
of $\cali$ that occur either in $C_i$ or in its OR handles, this
rules would have been canceled w.r.t. $\Pi$ as well, and thus
$\cali$ could not be stable. In situation (ii), there is some rule
that we would have canceled w.r.t. $P$ and has not been canceled for
$P_i$, i.e, there is a literal $\no \alpha$ which is false w.r.t.
$\cali$ and true w.r.t. $S_i$. But, if $\alpha \in \cali$ and
$\alpha$ occurs in $P_i$, then $\alpha \in S_i$ by hypothesis, and
thus (ii) cannot be the case as well.
\end{proof}

Once we get partial stable models of the composing cycles, we can
try to put them together in order to obtain stable models of the
whole program. Of course, we will try to obtain a stable model of
the overall program by taking one partial stable model for each
cycle, and considering their union. This however will work only if
the partial stable models assign truth values to atoms in a
compatible way.

\begin{definition}
\label{compatible} Consider a collection $\cals = S_1,\ldots,S_w$ of
partial stable models for $\Pi$, relative to its composing cycles
$C_1,\ldots,C_w$, each $S_i$ with base $\langle X_i, {X_i}^{-}
\rangle$. We say that $S_1,\ldots,S_w$ are {\em compatible} or,
equivalently, that $\cals$ is a {\em compatible set of partial
stable models} whenever the following conditions hold:
\begin{enumerate}
\item $\forall j,k \leq w$, $X_j \cap {X_k}^{-} = \emptyset$;
\item $\forall j \leq w$, $\forall A \in X_j$, $\exists h \neq j$
such that $A \in S_h$, and $A \not\in X_h$;
\item $\forall j \leq
w$, $\forall B \in {X_j}^{-}$, $\not\exists t \leq w$ such that $B
\in S_t$.
\end{enumerate}
\end{definition}

Condition (1) states that the bases of compatible partial stable
models cannot assign opposite truth values to any atom. Condition
(2) ensures that, if an atom $A$ is supposed to be true in the base
of some cycle $C_j$, it must be actually concluded true in some
other cycle $C_h$. Notice that ``concluded'' does not mean
``assumed'', and thus $A$ must occur in the partial stable model
$S_h$ of $C_h$, without being in its set of assumptions $X_h$.
Condition (3) ensures that, if an atom is supposed to be false in
the base of some cycle, it cannot be concluded true in any of the
other cycles.

The following result formally states the connection between the
stable models of $\Pi$, and the partial stable models of its cycles.

\begin{theorem}
\label{spartial} \label{stableiff} Let $\Pi$ be a program with
composing cycles $C_1,\ldots,C_w$ and  $\cali$ be a set of atoms.
$\cali$ is a stable model of $\Pi$ if and only if $I = {\bigcup}_{i
\leq w} S_i$ where each $S_i$ is a partial stable model for $C_i$
and ${\cals} = S_1,\ldots,S_w$ is a compatible set of partial stable
models.
\end{theorem}

\begin{proof}
Suppose that $\cali$ is a stable model for $\Pi$. Let $C_i$, $i \leq
w$, be any of the composing cycles of $\Pi$. Let $X_i = \cali \cap
Atoms(H_{C_i})$, which means that $X_i$ is the set of the atoms of
the handles of $C_i$ which are true w.r.t. $\cali$, and $X_i^{-} =
Atoms(H_{C_i}) \setminus X_i$. Let $S_i$ be the restriction of
$\cali$ to the atoms involved in the extended cycle $P_i = C_i +
Aux(C_i) + X_i$. By Lemma\rif{restriction} $S_i$ is a stable model
for $P_i$.  Then, it remains to prove that $S_{1}, \ldots, S_{w}$
form a compatible set of partial stable models. Condition 1 of
Definition\rif{compatible} holds because by construction we put in
$X_i$ all the atoms occurring in the handles of $C_i$ that are true
w.r.t. $\cali$: should they occur in the handles of some other cycle
$C_j$ they should be in $X_j$, and not in $X^{-}_j$. For condition
2, notice that atoms in $X_i$ do not occur in the head of the rules
of $C_i$. Since they occur in $\cali$, they must have been derived
by means of the rules of some other cycle $C_j$, and thus they occur
in $S_j$. For condition 3, it is sufficient to notice that we put in
the $X^{-}_j$'s atoms that are not in $\cali$, and consequently are
not in the $S_i$'s.

Vice versa, let us consider a compatible set ${\cals} = S_{1},
\ldots ,S_{w}$ of partial stable models for the cycles in $\Pi$.
Notice that $\Pi$ itself corresponds to the union of the cycles and
of their auxiliary rules, i.e., $\Pi = \bigcup_{i \leq w} C_i +
Aux(C_i)$.

Let us first show that $\cali = {\bigcup}_{i \leq w} S_i$ is a
stable model of the program ${\Pi}_L$ $=$ $\bigcup_{i \leq w} C_i +
Aux(C_i) + X_i$, which is a superset of $\Pi$. In fact, each $S_i$
satisfies the stability condition on the rules of the corresponding
extended cycle, and, since they form a compatible set, by conditions
(1) and (3) of Definition\rif{compatible} no atom which is in the
negative base of any of the $S_i$'s, is concluded true in some other
$S_j$. Therefore, $\cali$ is a stable model of ${\Pi}_L$.

In order to obtain $\Pi$ from ${\Pi}_L$, we have to remove the
positive bases of cycles, which are the unit rules corresponding to
the $X_i$'s. By condition (2) however, in a set of compatible
partial stable models every atom $A \in X_i$ is concluded true in
some $S_j$, $i \neq j$, i.e., in the partial stable model of some
other cycle. This implies that $\cali$ satisfies the stability
condition also after $X_i$'s have been removed: then, $\cali$ is a
stable model for $\Pi$.
\end{proof}

Each stable model $S$ of $\Pi$ corresponds to a different choice of
the $X_i$'s, i.e., of the active handles of cycles.

The above result is of theoretical interest, since it sheds light on
the connection between stable models of a program and stable models
of its sub-parts. It may also contribute to any approach to
modularity in software development under the stable model semantics.

From Theorem\rif{prel} and Theorem\rif{spartial} we can argue that
for checking whether a logic program has stable models (and,
possibly, for finding these models) one can do the following.

\begin{itemize}
\item[(i)] Divide program $\Pi$ into pieces, of the form $C_i + Aux(C_i)$, and check whether
every odd cycle has handles; if not, then the program is
inconsistent.

\item[(ii)] For every cycle $C_i$ with handles, find the sets $X_{i}$ that make the
subprogram $C_i + Aux(C_i)$ consistent, and find the stable models
$S_{C_i}$ of each $C_i + Aux(C_i) + X_{i}$. Notice that in the case
of unconstrained even cycles, $H_{C_i}$ is empty, and we have two
stable models, namely $M_{C_i}^1$ = Even\_atoms($C_i$) and
$M_{C_i}^2$ = Odd\_atoms($C_i$).

\item[(iii)] Check whether there exists a collection of $X_{i}$'s, one for each cycle,
such that the corresponding $S_{C_i}$'s form a compatible collection
of partial stable models for $\Pi$: in this case the program is
consistent, and its stable model(s) can be obtained as the union of
the $S_{C_i}$'s.
\end{itemize}

To show how the method works, consider for instance the following
program.

\st $
\begin{array}{l}
\li OC_1\\
q \ar \no q\\
\lili Aux(OC_1)\\
q \ar f\\
\li OC_2\\
p \ar \no p, \no f\\
\li EC_1\\
e \ar \no f\\
f \ar \no e
\end{array} $

\ni It can be seen as divided into the following parts, each one
corresponding to $C_i + Aux(C_i)$ for cycle $C_i$. The first part is
composed of odd cycle $OC_1$, with an auxiliary rule (OR handle):

\st $
\begin{array}{l}
q \ar \no q\\
q \ar f
\end{array} $

\ni The second part is composed of odd cycle $OC_2$, without
auxiliary rules but with an AND handle:

\st $
\begin{array}{l}
p \ar \no p, \no f
\end{array} $

\ni The third part is composed of the unconstrained even cycle
$EC_1$:

\st $
\begin{array}{l}
e \ar \no f\\
f \ar \no e
\end{array} $

\sni $OC_1$ in itself is inconsistent, but if we take $X_{OC_1} =
\{f\}$ (and $X^{-}_{OC_1} = \emptyset$) we get the partial stable
model $\{f, q\}$: the active OR handle forces $q$ to be true.
Similarly, if we take for $OC_2$ $X_{OC_2} = \{f\}$, we get the
partial stable model $\{f\}$: the active AND handle forces $p$ to be
false. Cycle $EC_1$ is consistent, with partial stable models
$\{e\}$ and $\{f\}$. If we now select the partial stable model
$\{f\}$, we get a compatible set of partial stable models thus
obtaining the stable model $\{f,q\}$ for the overall program.
Instead, the partial stable model $\{e\}$ for $EC_1$ does not serve
to the purpose of obtaining a stable model for the overall program,
since condition 2 of Definition\rif{compatible} is not fulfilled.
This in particular means that atom $f$, which is in the positive
base of both the odd cycles, is not concluded true in this partial
stable model. Therefore, the handles of the odd cycles are not
active and no overall consistency can be achieved.

Take now this very similar program, that can be divided into cycles
analogously.

\st $
\begin{array}{l}
\li OC_1\\
q \ar \no q\\
\lili Aux(OC_1)\\
q \ar f\\
\li OC_2\\
p \ar \no p, \no e\\
\li EC_1\\
e \ar \no f\\
f \ar \no e
\end{array} $

\ni The difference is that $OC_2$ has AND handle $\no e$ (instead of
$\no f$). With base $X_{OC_2} = \{e\}$ we get partial stable model
$\{e\}$ To fulfill condition 2 of Definition\rif{compatible}, we
should select partial stable model $\{e\}$ of $EC_1$. Unfortunately
however, since $OC_1$ is consistent only if we take $X_{OC_1} =
\{f\}$, we should at the same time choose the other partial stable
model $\{f\}$ of $EC_1$. Thus, no choice can be made for $EC_1$ so
as to make this program consistent.

With the aim of developing effective software engineering tools and
more efficient algorithms for computing stable models, syntactic
conditions for the existence of stable models are in order. In the
ongoing, we will use the above results as the basis for defining
necessary and sufficient syntactic conditions for consistency.

\section{Handle assignments and admissibility}

In previous sections we have discussed how to split a stable model
of a program into a compatible set of partial stable models of the
cycle. However, we have not formalized a method for selecting bases
for the cycles so as to ensure that all cycles have partial stable
models, and that they form a compatible collection. To this aim, in
this section we define syntactic condition that specify how active
handles affect consistency of extended cycles.

Assume that atom $\alpha$ occurs in some handle $\Delta$ of a cycle.
Then, it may possibly occur in the positive or negative base of that
cycle for forming a partial stable model that may be a part of a
compatible collection. In this case: if $\alpha$ occurs in the
positive base, then by condition 2 of Definition\rif{compatible}it
must be concluded true in some other cycle; if instead $\alpha$
occurs in the negative base, then by condition 3 of
Definition\rif{compatible} it must not be concluded true in any
other cycle. Notice that the cycles where it is possible to derive
an atom $\alpha$ are the cycles $\alpha$ is involved in, which are
the cycles the handle $\Delta$ \be comes from, \ee or equivalently
the \be source cycles \ee of the handle.

\begin{definition}
A handle $(\Delta: K)$ of cycle $C_1$, $\Delta = \alpha$ or
$\Delta = \no \alpha$ \be comes from  \ee source cycle $C_2$ if
$\alpha \in Composing\_atoms(C_2)$.
\end{definition}

Handles in $H_C$ are called the \be incoming handles \ee of $C$.
The same handle of a cycle $C$ may come from different cycles,
and may refer to different atoms of $C$.
For instance, in the program below we have:

\st $
\begin{array}{l}
\li OC_1\\
a \ar \no b, \no f\\
b \ar \no c\\
c \ar \no a, \no f\\
b \ar g\\
\li EC_1\\
f \ar \no g\\
g \ar \no f\\
\li EC_2\\
f \ar \no h\\
h \ar \no f
\end{array} $

handle $(\no f : AND)$ of $OC_1$ comes from both $EC_1$ and $EC_2$, and refers
to two different atoms in $OC_1$, namely $a$ and $c$; handle $(g : OR)$ of $OC_1$
comes from $EC_1$, and refers to atoms $b$.

The following definition completes the terminology by identifying
the atoms occurring in handles coming from $C$.

\begin{definition}
Given a cycle $C$, the set of the atoms involved in $C$ that occur
in the handles of some other cycle is defined as follows:\\
$Out\_handles(C) = \{\beta  \, | \\
\hspace{1cm} \beta \in Composing\_Atoms(C) \wedge \exists C_1 \neq C
\mbox{\,such that\,} \beta \in Atoms(H_{C_1})\}$
\end{definition}

In the above program for instance, we have $Out\_handles(EC_1)$ $=$
$\{f,g\}$ and $Out\_handles(EC_2)$ $=$ $\{f\}$. Notice that,
according to the definition, $h \not\in Out\_handles(EC_2)$, because
$h$ does not occur in any other cycle.

For an handle to be active w.r.t. a set of atoms, we must have the
following. (i) If the corresponding atom $\alpha$ is required to be
true, then it must be concluded true (by means of a supporting rule)
in {\em at least one} of the cycles the handle comes from, which
implies $\alpha$ to be concluded true in {\em all} the
\textit{extended} cycles it is involved into: in fact, the rule that
makes $\alpha$ true is an auxiliary rule for all these cycles. (ii)
If the corresponding atom $\alpha$ is required to be false, then it
must be concluded false in {\em all} the (extended) cycles it comes
from.

This is illustrated by the following example:

\st $
\begin{array}{l}
\li OC_1\\
p \ar \no p, \no c\\
\li OC_2\\
c \ar \no d\\
d \ar \no e\\
e \ar \no c, f\\
\li EC_1\\
f \ar \no g\\
g \ar \no f\\
\li EC_2\\
f \ar \no h\\
h \ar \no f
\end{array} $

The extended cycles are:

\st $
\begin{array}{l}
\li OC_1\\
p \ar \no p, \no c
\end{array} $

\ni with no auxiliary rules, $Out\_handles(EC_1) = \emptyset$,
$H_{OC_1}$ $=$ $\{\no c\ :\ AND\ :\ p\}$, $Atoms(H_{OC_1}) = \{c\}$
and unique partial stable model $\{c\}$ obtained by choosing
$X_{OC_1} = \{c\}$;

\st $
\begin{array}{l}
\li OC_2\\
c \ar \no d\\
d \ar \no e\\
e \ar \no c, f
\end{array} $

\ni with no auxiliary rules, $Out\_handles(OC_2) = \{c\}$,
$H_{OC_2}$ $=$ $\{f\ :\ AND\ :\ e\}$, $Atoms(H_{OC_2}) = \{f\}$ and
unique partial stable model $\{d\}$ obtained by choosing positive
base $X_{OC_2} = \emptyset$, $X^{-}_{OC_1} = \{f\}$;

\st $
\begin{array}{l}
\li EC_1 + Aux(EC_1)\\
f \ar \no g\\
g \ar \no f\\
f \ar \no h
\end{array} $

\ni with $Out\_handles(EC_1) = \{f\}$, $H_{EC_1}$ $=$ $\{\no h\ :\
OR\ :\ f\}$, $Atoms(H_{EC_1}) = \{h\}$. Notice that if we take
$X_{EC_1} = \{h\}$ this makes $f$ not derivable via the auxiliary
rule, while however $f$ is still derivable via the corresponding
rule involved in the cycle. Then, if we consider the two partial
stable models $\{f\}$ and $\{g\}$ of $EC_1$, we see that: the former
one can be obtained either by choosing either $X_{EC_1} = \emptyset$
or $X_{EC_1} = \{h\}$; the latter one instead requires $X_{EC_1} =
\{h\}$, so as to allow $f$ to be false.

\st $
\begin{array}{l}
\li EC_2  + Aux(EC_2)\\
f \ar \no h\\
h \ar \no f\\
f \ar \no g
\end{array} $

\ni with $Out\_handles(EC_2) = \{f\}$, $H_{EC_2}$ $=$ $\{\no g\ :\
OR\ :\ f\}$, $Atoms(H_{EC_2}) = \{g\}$ and two partial stable models
$\{f\}$ and $\{h\}$. The former one can be obtained either by
choosing $X_{EC_2} = \emptyset$ or, also, $X_{EC_1} = \{g\}$. The
latter one requires $X_{EC_2} = \{g\}$, so as to allow $f$ to be
false.

Unfortunately, the overall program turns out to have no stable
model, because: for obtaining the partial stable model of $OC_2$,
$f$ must be concluded false so to make the unique AND handle active.
Both $EC_1$ and $EC_2$ actually admit a partial stable model where
$f$ is false. Thus, for the fragment $EC_1 + EC_2 + OC_2$ we may
construct the unique wider partial stable model $\{g, h, d\}$.
However, this fails to make the handle of $OC_1$ active, and
therefore a stable model for the program cannot be obtained.

Assume to replace $OC_1$ with ${OC'}_1$

\st $
\begin{array}{l}
\li {OC'}_1\\
p \ar \no p, \no d
\end{array} $

\ni where $H_{{OC'}_1}$ $=$ $\{\no d\ :\ AND\ :\ p\}$,
$Atoms(H_{{OC'}_1}) = \{d\}$ and there is a unique partial stable
model $\{d\}$ obtained by choosing positive base $X_{OC_1} = \{d\}$.
In this case, $\{g, h, d\}$ would be a stable model for the overall
program.

Assume to add the cycles:

\st $
\begin{array}{l}
\li {OC}_3\\
q \ar \no q, d\\
\\
\li {OC}_4\\
r \ar \no r\\
r \ar \no d.
\end{array} $

The resulting program cannot be consistent. On the one hand in fact,
${OC'}_1$ and $OC_3$ have unique AND handles $\no d$ and $d$
respectively, that cannot be active at the same time. On the other
hand, ${OC'}_1$ has an AND handle $\no d$ while $OC_4$ has an OR
handle $\no d$, and also in this case these handles cannot be active
at the same time.

Below we establish the formal foundations of the kind of reasoning
that we have informally proposed up to now. Some more definitions
about handles are needed.

\begin{definition}
The handles $(\Delta : AND)$ and $(\Delta : OR)$ are called \be
opposite handles. \ee Given a handle $h$, we will indicate its
opposite handle with $h^{-}$.
\end{definition}

\begin{definition}
The handles $(\Delta_1 : K)$ and $(\Delta_2 : K)$ are called \be
contrary handles \ee if $\Delta_1 = \alpha$ and $\Delta_2 = \no
\alpha$. Given a handle $h$, we will indicate its contrary handle
with $h^n$.
\end{definition}

Whenever either contrary or opposite pairs of handles occur in a
program, even for different $\beta$'s, if one is active w.r.t. a
given set of atoms then the other one is not active, and vice versa.

\begin{definition}
The handles $(\Delta_1 : K1)$ and $(\Delta_2 : K2)$ are called \be
sibling handles \ee if $K1 \neq K2$, and $\Delta_1 = \alpha$ and
$\Delta_2 = \no \alpha$. Given a handle $h$, we will indicate its
sibling handle with $h^s$.
\end{definition}

Whenever sibling pairs of handles occur in a program, even for
different $\beta$'s, if one is active if one is active w.r.t. a
given set of atoms then the other one is active as well.

Taken for instance atom $\alpha$, we have:

\begin{itemize}
\item
$(\alpha : AND)$ and $(\alpha : OR)$ are opposite handles;
\item
$(\no \alpha : AND)$ and $(\no \alpha : OR)$ are opposite handles;
\item
$(\alpha : AND)$ and $(\no \alpha : AND)$ are contrary handles;
\item
$(\alpha : OR)$ and $(\no \alpha : OR)$ are contrary handles;
\item
$(\alpha : OR)$ and $(\no \alpha : AND)$ are sibling handles;
\item
$(\alpha : AND)$ and $(\no \alpha : OR)$ are sibling handles;
\end{itemize}

Finally, we introduce the definition of {\em handle assignment},
which is a consistent hypothesis on (some of) the handles of a cycle
$C$. Namely, it is a quadruple composed of the following sets.
$IN^A_C$ contains the incoming handles which are assumed to be
active. From $IN^A_C$ one can immediately derive a corresponding
assumption on $X_C$. In particular, \(X_C = ActivationAt_C(IN^A_C)
\), i.e. it is exactly the set of the atoms that make the handles in
$IN^A_C$ active. Vice versa, $IN^N_C$ contains the incoming handles
which are assumed to be not active. Handles of $C$ which are not in
$IN^A_C \cup IN^N_C$ can be either active or not active, but their
status is either unknown or irrelevant in the context where the
handle assignment is used.

$OUT_C^{+}$ is the set of out-handles which are required to be
concluded true. This in order to make some handle of some other
cycle active, as we have seen in the example above. Similarly,
$OUT_C^{-}$ is the set of the out-handles which are required to be
concluded false, for the same reason. Of course, the $OUT_C$'s must
be disjoint, since no atom can be required to be simultaneously true
and false.

\begin{definition}
A {\em basic handle assignment} to (or for) cycle $C$ is a quadruple
of sets
\[\langle IN^A_C,IN^N_C,OUT_C^{+},OUT_C^{-}\rangle\]
where the (possibly empty) composing sets are such that:\\
$IN^A_C \cup IN^N_C \subseteq H_C$;\\
$IN^A_C \cap IN^N_C = \emptyset$;\\
neither $IN^A_C$ and $IN^N_C$ contain pairs of either opposite or
contrary handles;\\
$OUT_C^{+} \cup OUT_C^{-} \subseteq Out\_handles(C)$;\\
$OUT_C^{+} \cap OUT_C^{-} = \emptyset$.
\end{definition}

For short, when talking of both $IN^A_C$ and $IN^N_C$ we will say
\as the $IN_C$'s''. A handle assignment will be called {\em trivial}
(resp. non-trivial) if $OUT_C^{+} = OUT_C^{-} = \emptyset$, i.e.,
whenever there is no requirement on the out-handles of $C$.

If $IN^A_C$ is empty, there are two possible situations. (i) $H_C =
\emptyset$, i.e., the cycle is unconstrained. (ii) $H_C \neq
\emptyset$ but no active incoming handle is assumed: in this case,
we say that the cycle is \be actually unconstrained \ee w.r.t. this
handle assignment. A handle assignment will be called \be effective
\ee (w.r.t. non-effective) whenever $IN^A_C \neq \emptyset$.

We have to cope with the relationship between opposite, contrary, and sibling handles,
whenever they should occur in the same cycle $C$.

\begin{definition}
Let: $h$ and $h^{-}$ be a pair of opposite handles; $h$ and $h^n$ be
a pair of contrary handles; and $h$ and $h^s$ be a pair of sibling
handles. A {\em complete handle assignment}, or simply a {\em handle
assignment}, to cycle $C$ is a basic handle assignment to $C$ where,
for each pair of
opposite, contrary or sibling handles the occur in $C$, the following conditions hold:\\
$h \in IN^A_C$ if and only if $h^{-} \in IN^N_C$;\\
$h \in IN^A_C$ if and only if $h^n \in IN^N_C$;\\
then either $h, h^s \in IN^A_C$ and $h, h^s \not\in IN^N_C$ or $h,
h^s \in IN^N_C$ and $h, h^s \not\in IN^A_C$.
\end{definition}

A basic handle assignment can be \textit{completed}, i.e., turned into a complete
handle assignment, by an obvious update of the $IN_C$'s.

What the definition does not state yet is that $IN_C$'s and the
$OUT_C$'s should be compatible, in the sense that the handles in
$IN^A_C$ and $IN^N_C$ being active should not prevent the
out-handles in $OUT_C$'s from being true/false as required. Consider
for instance the following extended cycle $OC$, which is meant to be
a fragment of a wider program:

\st $
\begin{array}{l}
\li OC\\
a \ar \no b, f\\
b \ar \no c\\
c \ar \no a\\
b \ar \no e
\end{array} $

\ni where $H_{OC} = \{(\no e\ :\ OR\ :\ b), (f\ :\ AND\ :\ a)\}$.\\
Let us assume that $Out\_handles(OC) = \{a, b\}$: this means, we
assume that these are the atoms involved in $OC$ that occur in the
handles of some other cycle. Now take a handle assignment with the
following components. $IN^A_{OC} = \{(f\ :\ AND\ :\ a)\}$ which
means that we assume this handle to be active, i.e., we assume $f$
to be false. $IN^N_{OC} = \{(\no e\ :\ OR\ :\ b)\}$, which means
that we assume this handle to be not active, i.e., we assume $\no e$
to be false. Finally, $OUT_{OC}^{+} = \{b\}$, and $OUT_{OC}^{-} =
\{c\}$. For this handle assignment, the requirements about the
out-handles are not compatible with the assumptions on the incoming
handles. In fact, if $f$ is assumed to be false, then $a$ is
concluded false, and consequently $c$ is concluded true and $b$
false (since moreover the OR handle $\no e$ of $b$ is in
$IN^N_{OC}$, and thus is assumed to be not active). Notice that even
with $IN^N_{OC} = \emptyset$ (i.e., with no knowledge about handle
$(\no e\ :\ OR\ :\ b)$) still with the given assumptions about the
incoming handles we cannot assume to meet the requirements for
$OUT_{OC}^{+}$ and $OUT_{OC}^{-}$. Instead, with the same
$IN_{OC}$'s, and with $OUT_{OC}^{+} = \{c\}$ and $OUT_{OC}^{-} =
\emptyset$ we obtain an handle assignment where the assumptions are
compatible with the requirements. Notice also that $OUT_{OC}^{-} =
\emptyset$ does not mean that no out-handle is {\em allowed} to be
false, rather it means that no out-handled is {\em required} to be
false. Then, provided that the requirements in $OUT_{OC}^{+}$ and
$OUT_{OC}^{-}$ are met, the remaining out-handles can take any truth
value. Notice finally that if we let $IN^A_{OC} = IN^N_{OC} =
\emptyset$, then the extended cycle is inconsistent.

Clearly, a definition of the $IN_{OC}$'s that makes the
corresponding program fragment $C + Aux(C) + ActivationAt_C(IN^A_C)$
inconsistent is useless for obtaining stable models of the overall
program. In fact, we are interested in handle assignments where the
$IN_{OC}$'s correspond to an assumption on the incoming handles (and
thus on $X_C = ActivationAt_C(IN^A_C)$) such that: the resulting
program fragment $C + Aux(C) + X_C$ is consistent, and  the
requirements established in $OUT_{OC}^{+}$ and $OUT_{OC}^{-}$ are
met. This means that in some partial stable model of the program
fragment all atoms in $OUT_{OC}^{+}$ are deemed true, and all atoms
in $OUT_{OC}^{-}$ are deemed false.

This is formalized in the following:

\begin{definition}
\label{admissible_ass} A handle assignment $HA = \langle IN^A_C,
IN^N_C, OUT_C^{+},OUT_C^{-}\rangle$ to a cycle $C$ is {\em
admissible} if and only if the program $C + Aux(C) +
ActivationAt_C(IN^A_C)$ is consistent, and for some stable model
$S^{IN^A_C}$ of this program, $OUT_C^{+} \subseteq S^{IN^A_C}$ and
$OUT_C^{-} \cap S^{IN^A_C}  = \emptyset$. We say that $S^{IN^A_C}$
{\em corresponds} to $HA$.
\end{definition}

According to Definition\rif{partialstable}, each stable model $S^{IN^A_C}$ is a
partial stable model of $\Pi$ relative to $C$, that can be used
for building a stable model of the whole program. At least some of
these partial stable models \be correspond \ee to the given handle
assignment, in the sense that they are consistent with the choice
of active handles that the assignment represents.

\begin{proposition}
A non-effective handle assignment cannot be admissible for an odd
cycle.
\end{proposition}
\begin{proof}
A non-effective handle assignment provides an empty set of active
handles to the cycle, which then is either unconstrained (no
effective handle assignment exists because there are no handles) or
is actually unconstrained (no handle is made active by this
assignment). An unconstrained odd cycle is inconsistent. By
Theorem\rif{prel}, an actually unconstrained odd cycle is
inconsistent as well, since it has no active handle. Then, by
Definition\rif{admissible_ass} the assignment is not admissible.
\end{proof}

and also that:

\begin{proposition}
A non-effective handle assignment is admissible for an even cycle
$C$ if and only if either $OUT_C^{+} \subseteq Even\_atoms(C)$ or
$OUT_C^{+} \subseteq Odd\_atoms(C)$.
\end{proposition}
\begin{proof}
If cycle $C$ is even, and it is either unconstrained or actually
unconstrained, then the program fragment $C + Aux(C) +
ActivationAt_C(IN^A_C)$ has two stable models, $S_1$ coinciding with
$Even\_atoms(C)$, and $S_2$ coinciding with $Odd\_atoms(C)$. This
because the handles either do not exist or are not active, and thus
do not affect the stable models. By Definition\rif{admissible_ass}
the assignment is effective if and only if $OUT_C^{+} \subseteq S_1$
or $OUT_C^{+} \subseteq S_2$.
\end{proof}

Observe that whenever a handle assignment is effective the
corresponding program fragment is locally stratified, and thus,
according to \cite{teodor:survey}, has a unique stable model that
coincides with its well-founded model. It may also be observed that
a trivial handle assignment, which does not state requirements on
the out-handles, is always admissible for an even cycle, and it is
admissible for an odd cycle only if it is effective (otherwise as
seen before the cycle is inconsistent).

The admissibility of a non-trivial effective handle assignment for
cycle $C$ can be checked syntactically, by means of the
criterion that we state below. The advantage of this check is that it does not require
to compute the well-founded model of $C + Aux(C) +
ActivationAt_C(IN^A_C)$, but it just looks at the rules of $C$.
Although the syntactic formulation may seem somewhat complex, it
simply states in which cases an atom in $OUT_C^{+}$, which is
required to be concluded true w.r.t. the given handle assignment
(or, conversely, an atom in $OUT_C^{-}$ which is required to be
concluded false), is actually allowed to take the specified truth
value without raising inconsistencies. Notice that $OUT_C^{+}$ and
$OUT_C^{-}$ must be mutually coherent, in the sense that truth of
an atom in $OUT_C^{+}$ cannot rely on truth of an atom in
$OUT_C^{-}$ (that instead is required to be concluded false), and
vice versa.

\begin{proposition}
\label{admissible}
A non-trivial effective handle assignment $\langle IN^A_C$, $IN^N_C$, $OUT_C^{+}$,
$OUT_C^{-}\rangle$ to cycle $C$ is admissible if and only if for
every $\lambda_i \in OUT_C^{+}$ the following condition (1) holds,
and for every $\lambda_k \in OUT_C^{-}$ the following condition
(2) holds.
\begin{enumerate}
\item
Condition 1.

\begin{enumerate}
\item
Either there exists OR handle $h_o$ for $\lambda_i$, $h_o \in
IN^A_C$ or
\item
for every AND handle $h_a$ for $\lambda_i$, $h_a \in IN^N_C$ and\\
$\lambda_{i+1} \not\in OUT_C^{+}$, and \\
condition (2) holds for $\lambda_{i+1}$.
\end{enumerate}
\item
Condition 2.

\begin{enumerate}
\item
For every OR handle $h_o$ for $\lambda$, $h_o \in IN^N_C$, and
\item
either there exists AND handle $h_a$ for $\lambda$ such that $h_a
\in IN^A_C$,
or\\
$\lambda_{k+1} \not\in OUT_C^{-}$, and
condition (1) holds for $\lambda_{k+1}$.
\end{enumerate}
\end{enumerate}
\end{proposition}

\begin{proof}
Let us first notice that the set of rules with head $\lambda_{i}$
in $C + Aux(C) + ActivationAt_C(IN^A_C)$ consists of rule
$\lambda_i \ar \no \lambda_{i+1}, \Delta_i$ in $C$, and
possibly, of one or more rules in $Aux(C)$.
In fact, by the definition of canonical program, atoms in $IN^A_C$ do not
occur in $C$, and thus $\lambda_i$ cannot belong to $ActivationAt_C(IN^A_C))$.

Consider an atom $\lambda_i \in OUT_C^{+}$, that we want to be
concluded true in the partial stable model of $C$ which corresponds
to the given handle assignment. For $\lambda_i$ to be concluded
true, there must be a rule whose conditions are true w.r.t the
handle assignment.

One possibility, formalized in Condition 1.(a),
is that there exists an OR handle $h_o$ for $\lambda_i$,
$h_o \in IN^A_C$.
That is, there is an auxiliary rule with head $\lambda_i$, and
condition true w.r.t. the handle assignment.

\ni
Otherwise, as formalized in Condition 1.(b)
we have to consider the rule of cycle $C$:

\st $
\begin{array}{l}
\lambda_i \ar \no \lambda_{i+1}, \Delta_i\\
\end{array} $

\st and check that all the conditions are guaranteed to be true by
the handle assignment. First of all it must be $(\Delta_k : AND :
\lambda_k) \in IN^N_C$ i.e., in the given handle assignment the AND
handle referring to $\lambda_i$ must be not active, because an
active AND handle would make the head of the rule false. Second,
$\no \lambda_{i+1}$ must be true: this on the one hand requires
$\lambda_{i+1} \not\in OUT_C^{+}$, that would be a contradiction; on
the other hand, requires $\lambda_{i+1}$ to be concluded false. To
this aim, condition (2), discussed below, must hold for
$\lambda_{i+1}$.

\noindent Consider now an atom $\lambda_k \in OC_C^{-}$, that
we want to be false the partial stable model of $C$, which
corresponds to the given assignment: there must \be not \ee be a
rule for $\lambda_k$ whose conditions all true
w.r.t. the given assignment.

First, as formalized
in Condition 2.(a), we must have any OR handle $h_o$
for $\lambda_k$ in $IN^N_C$. Otherwise, $\lambda_k$ would be necessarily concluded
true, being the head of an auxiliary rule with a true body.

\noindent
Second, as formalized in Condition 2.(b), we also have to consider the rule of cycle $C$

\st $
\begin{array}{l}
\lambda_k \ar \no \lambda_{k+1}, \Delta_k\\
\end{array} $

\st and check that one of its two conditions is
false w.r.t. the handle assignment. A first case is that
$(\Delta_k : AND : \lambda_k) \in IN^A_C$, which means that the AND
handle referring to $\lambda_k$ is supposed to be active, i.e.,
false. Otherwise, $\no \lambda_{k+1}$ must be
false, i.e., $\lambda_{k+1}$ must be true. To this
aim, provided that $\lambda_{k+1} \not\in OUT_C^{-}$ (that would
be a contradiction), condition (1) must hold for $\lambda_{k+1}$.
\end{proof}

The fact that Conditions 1 and 2 refer to each other is not
surprising, since they are to be applied to cycles. Consider for
instance the following cycle:

\st $
\begin{array}{l}
e \ar \no f\\
f \ar \no g\\
g \ar \no e\\
g \ar h
\end{array} $

\ni The handle assignment $\langle
\{(h:OR)\},\emptyset,\{g\},\emptyset \rangle$ is admissible by
Proposition\rif{admissible}, since, according to Condition 1.(a),
there exists an auxiliary rule with head $g$ and body in $IN^A_C$.
Also $\langle \{(h:OR)\},\emptyset,\{g,e\},\emptyset \rangle$ is
admissible, because: $g$ is as above; there is no OR handle for $e$,
thus Condition 1.(a) cannot be applied, but, considering rule $e \ar
\no f$ (Condition 1.(b)), it is easy to see that Condition 2 holds
of $f$, since there is no OR handle for $f$, and we have just shown
that Condition 1 holds of $g$. Instead, $\langle
\{(h:OR)\},\emptyset,\{g\},\{e\} \rangle$ is not admissible, because
Condition 2 does not hold of $e$.

It is important to notice that it is possible to determine
admissible handle assignments from a partially specified one. An
obvious way of doing that is: guessing the missing sets, and
checking whether the resulting handle assignment is admissible.
There is however a much easier way by exploiting the definitions.

Namely, for given $IN_C$'s it is easy to find the maximal values for
$OUT_C^{+}$ and $OUT_C^{-}$ that form an admissible handle
assignment. If $IN^A_C$ is empty, then they correspond to the stable
models (if any) of the cycle taken by itself (without the auxiliary
rules, since an empty $IN^A_C$ means that no OR handle is active).
If $IN^A_C$ is not empty, then by asserting the atoms in
$ActivationAt_C(IN^A_C)$ as facts one computes the (unique) stable
model of the extended cycle, and thus the maximal values for
$OUT_C^{+}$ and $OUT_C^{-}$. These maximal values are determined by
assuming all handles not belonging to the $IN_C$'s to be not active.

Vice versa, given $OUT_C^{+}$ and $OUT_C^{-}$, and unknown or partially defined $IN_C$'s,
the conditions stated in Proposition\rif{admissible}
can be used
for determining the subsets of $H_C$ (incoming handles) that
form admissible handle assignments.

Consider for instance the extended cycle:

\st $
\begin{array}{l}
e \ar \no f, \no r\\
f \ar \no g\\
g \ar \no e\\
g \ar v\\
g \ar h\\
e \ar s\\
e \ar \no h
\end{array} $

and let $OUT_C^{+} = \{g\}$ and $OUT_C^{-} = \emptyset$. Then, for
forming an admissible handle assignment we have three possibilities.

First, by Condition 1.(a) of Proposition\rif{admissible},
we can exploit the auxiliary rule $g \ar v$, i.e. the handle $(v : OR)$,
and let $IN^{A1}_C = \{(v : OR)\}$, and $IN^{N1}_C = \emptyset$.

Second, again by Condition 1.(a) of Proposition\rif{admissible}, we
can exploit the other auxiliary rule $g \ar h$, i.e. the handle $(h
: OR)$, and let $IN^{A2}_C = \{(h : OR)\}$. This implies to insert
into $IN^{N2}_C$ the contrary and opposite handles, since they both
occur in $C$, i.e. $IN^{N2}_C = \{(h : AND), (\no h: OR)\}$.

Third, we can exploit condition 1.(b), and consider rule with head
$g$ in the cycle, i.e. $g \ar \no e$, and verify Condition 2 for
$e$, that must be false. For checking Condition 2.(a), we have to
consider both the OR handles for $e$, i.e. handle $(\no h: OR)$ and
handle $(s : OR)$, that must be included in $IN^{N2}_C$, i.e.,
$IN^{N3}_C = \{(\no h : OR), (s : OR)\}$. For checking Condition
2.(b) we have to consider rule $e \ar \no f, \no r$. Since we want
$g$ to be true, this implies $f$ to be false, which means that for
getting $e$ false as well, we have to add its AND handle $(\no r :
AND)$ to $IN^{N3}_C$. I.e., finally we get $IN^{N3}_C = \{(\no h:
OR), (s:OR), (\no r : AND)\}$. This leads to add the opposite and
contrary handles which occur in $C$ to $IN^{A2}_C$, thus letting:
$IN^{A3}_C = \{(h: OR)\}$.

We may notice that $IN^{A2}_C = IN^{A3}_C$ but $IN^{N2}_C \subseteq
IN^{N3}_C$. Both choices form an admissible handle assignment,
although the first one is more restricted. It turns out in fact
that, in the above cycle, for building handle assignments where
$OUT_C^{+} = \{g\}$ and $OUT_C^{-} = \emptyset$, the handle $(\no r
: AND)$ is actually irrelevant. This explains why the definition of
handle assignment does not enforce one to set \be all \ee the
handles of the cycle as active/not active. We can introduce the
following definition:

\begin{definition}
An admissible handle assignment $\langle IN^A_C, IN^N_C,OUT_C^{+}, OUT_C^{-}\rangle$ is
\be minimal \ee if there is no other sets $IN^{A'}_C \subset IN^A_C$ and $IN^{N'}_C \subset IN^N_C$ such that
$\langle IN^{A'}_C, IN^{N'}_C, OUT_C^{+}, OUT_C^{-}\rangle$ is still admissible.
\end{definition}

As we have just seen, there can be alternative minimal sets of
incoming active handles for the same out-handles. However, there may
also be the case there is none. There is for instance no possibility
for $OUT_C^{+} = \{g,f\}$, i.e., no choice for the $IN_C$'s can
produce a partial stable model where both $g$ and $f$ are true.

\section{Cycle Graph and support sets}

In previous sections we have proved that a stable model of a program
can be obtained as the union of a compatible collection of partial
stable models of the composing cycles. Partial stable models of a
cycle are obtained by considering the corresponding extended cycle
as a program, and making assumptions about its handles. We have
discussed how to study consistency of extended cycles by means of
the notion of handle assignment. In this section we introduce the
Cycle Graph of a program, that represent cycles and handles. In the
rest of the paper we show that the concepts and principles that we
have previously introduced allow us to define syntactic conditions
for consistency on the Cycle Graph.

\begin{definition}
Given program $\Pi$, the {\em Cycle Graph} $CG_{\Pi}$, is a directed
graph defined as follows:
\begin{itemize}
\item {\bf Vertices.} One vertex for each of cycles
$C_1,\ldots,C_w$ that occur in $\Pi$. Vertices corresponding to even
cycles are labeled as $EC_i$'s while those corresponding to odd
cycles are labeled as $OC_j$'s. \item {\bf Edges.} An edge
$(C_j,C_i)$ marked with $(\Delta : K : \lambda)$ for each handle
$(\Delta : K : \lambda) \in H_{C_i}$ of cycle $C_i$, that comes from
$C_j$.
\end{itemize}
\end{definition}

Each marked edge will be denoted by $(C_j,C_i | \Delta : K :
\lambda)$, where either $(C_j$ or $C_i$ or $\lambda)$ will be
omitted whenever they are clear from the context, and we may write
for short $(C_j,C_i | h)$, $h$ standing for a handle that is either
clear from the context or does not matter in that point. An edge on
the $CG$ connects the cycle a handle comes from to the cycle to
which the handle belongs.

Take for instance the following program $\pi_0$.

\st $
\begin{array}{l}
\li OC_1\\
p \ar \no p,\no c\\
\lili Aux(OC_1)\\
p \ar \no b\\
\li OC_2\\
q \ar \no q\\
\lili Aux(OC_2)\\
q \ar a\\
q \ar \no e\\
\li OC_3\\
r \ar \no r, \no e\\
\li EC_1\\
c \ar \no d\\
d \ar \no c\\
\li EC_2\\
a \ar \no b\\
b \ar \no a\\
\li EC_3\\
e \ar \no f\\
f \ar \no e
\end{array} $

Its cycle graph $CG_{\pi_0}$ is shown in Figure 1.

\begin{figure}[htbp]
\label{fig:CGEX}
    \begin{center}
    \epsfig{file=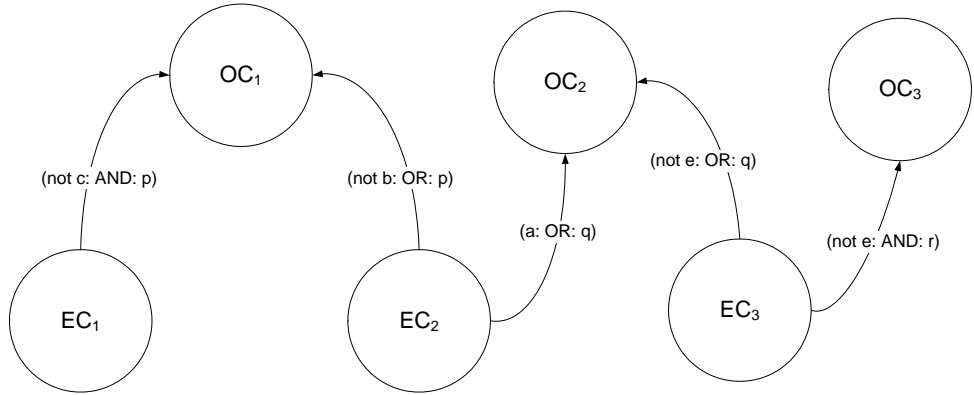, scale=0.4}
    \end{center}
\caption{The Cycle Graph of $\pi_0$.}
\end{figure}

The Cycle Graph of a program directly represents cycles, that
correspond to its vertices. It also indirectly represents extended
cycles, since its edges are marked by the handles. Paths on the
Cycle Graph graph represent direct or indirect connections between
cycles through the handles. In order to relate admissible handle
assignments for the cycles of $\Pi$ to subgraphs of its cycle graph
$CG_{\Pi}$ we introduce the following definitions.

\begin{definition}
\label{CGSS} Given program $\Pi$, let a \be CG support set \ee be a
pair $S = \langle ACT^{+}, ACT^{-} \rangle$ of subsets of the
handles marking the edges of $CG_{\Pi}$ (represented in the form
$(\Delta : K)$ with $K = AND/OR$) such that the following conditions hold: \\
(i) $ ACT^{+} \cap ACT^{-} = \emptyset$.\\
(ii) neither $ ACT^{+}$ nor $ACT^{-}$ contain a pair of either opposite or contrary handles.\\
(iii) if two opposite handles $h$ and $h^{-}$ both occur on the $CG$,
then $ACT^{+}$ contains handle $h$ if and only if $ ACT^{-}$ contains its opposite handle $h^{-}$.\\
(iv) if two contrary handles $h$ and $h^n$ both occur on the $CG$, then $ACT^{+}$
contains handle $h$ if and only if $ ACT^{-}$ contains its contrary handle $h^n$.\\
(v) if two sibling handles $h$ and $h^s$ both occur on the $CG$, then either $h, h^s \in ACT^{+}$ and
$h, h^s \not\in ACT^{-}$, or vice versa $h, h^s \in ACT^{-}$ and $h, h^s \not\in ACT^{+}$
\end{definition}

Given $S$, we will indicate its two components with $ACT^{+}(S)$ and
$ACT^{-}(S)$. For the sake of readability we introduce some
simplifying assumptions.

\begin{itemize}
\item
Given handle $h = (\Delta : K : \lambda)$, by
$ACT^{+}(S) \cup \{h\}$ (resp. $ACT^{-}(S) \cup \{h\}$) we mean
$ACT^{+}(S) \cup \{(\Delta : K)\}$ (resp. $ACT^{+}(S) \cup \{(\Delta : K)$\}).
\item
Given handle $h \in ACT^{+}(S)$ (resp. $h \in ACT^{-}(S)$)
of the form $(\Delta : K)$,  by $IN^A_C \cup \{h\}$ (resp. $IN^N_C \cup \{h\}$)
we mean: to identify the set $H = \{(\Delta : K : \lambda) \in H_C\}$ and perform $IN^A_C \cup H$
(resp. $IN^N_C \cup H$).
\item
By $H_C \cap ACT^{+}(S)$ (resp. $H_C \cap ACT^{+}(S)$)
we mean $\{(\Delta : K : \lambda) \in H_C | (\Delta: K) \in ACT^{+}(S)\}$ (resp. $(\Delta: K) \in ACT^{+}(S)$).
\end{itemize}

As stated in Theorem\rif{prel}, we have to restrict the attention on
CG support sets including at least one active handle for each odd
cycle. Then, according to Theorem\rif{spartial}, we have to check
that the assumptions on the handles are mutually coherent, and are
sufficient for ensuring consistency.

\begin{definition}
\label{potadequate} A CG support set $S$ is {\em potentially
adequate} if for every odd cycle $C$ in $\Pi$ there exists a handle
$h \in H_C$ such that $h \in ACT^{+}(S)$.
\end{definition}

A CG support set $S$ induces a set of handle assignments, one for each of
the cycles $\{C_1,\ldots,C_w \}$ occurring in $\Pi$.

The induced assignments are obtained on the basis of the following observations:

\begin{itemize}
\item
Each handle in $h \in ACT^{+}(S)$ is supposed to be active, and
therefore it must be active for each cycle $C_i$ such that $h \in H_{C_i}$.
\item
Each handle in $h \in ACT^{-}(S)$ is supposed to be not active, and
therefore it must be not active for each of cycle $C_j$ such that $h \in H_{C_j}$.
\item
If a handle $h$ in $S$ requires, in order to be active/not active,
an atom $\beta$ to be false, then it must be concluded false \be in
all the extended cycles of the program \ee $h$ comes from.
\item
If a handle $h$ in $S$ requires, in order to be active/not active,
an atom $\beta$ to be true, then it must be concluded true \be in
all the extended cycles of the program \ee $h$ comes from. This
point deserves some comment, since one usually assumes that it
suffices to conclude $\beta$ true \be somewhere \ee in the program.
Consider however that any rule $\beta \ar Body$ that allows $\beta$
to be concluded true in some cycle is an auxiliary rule to all the
other cycles $\beta$ is involved into. This is why $\beta$ is
concluded true \be everywhere it occurs. \ee This is the mechanism
for selecting partial stable models of the cycles that agree on
shared atoms, in order to assemble stable models of the overall
program.
\end{itemize}

\begin{definition}
\label{induced}
Let $S = \langle ACT^{+}, ACT^{-} \rangle$  be a GG support set which is potentially adequate.
For each cycle $C_k$ occurring in $\Pi$, $k \leq w$,
the (possibly empty) handle assignment induced by this set is determined
as follows.
\begin{enumerate}
\item
Let $IN^A_{C_k}$ be $H_{C_k} \cap ACT^{+}(S)$.
\item
Let $IN^N_{C_k}$ be $H_{C_k} \cap ACT^{-}(S)$.
\item
Let $OUT_{C_k}^{+}$ be the (possibly empty) set of all atoms $\beta \in Out\_handles(C_k)$
such that there is a handle $h \in ACT^{+}(S)$
either of the form $(\beta : OR)$ or $(\no \beta : AND)$.
\item
Let $OUT_{C_k}^{-}$ be the (possibly empty) set of all atoms $\alpha \in Out\_handles(C_k)$
such that there is a handle $h \in ACT^{-}(S)$
either of the form  $(\alpha : AND)$ or $(\no \alpha : OR)$.
\item
Verify that $OUT_{C_k}^{-} \cap OUT_{C_k}^{+} = \emptyset$.
\end{enumerate}
If this is the case for each $C_k$, then $S$
actually induces a set of handle assignments, and
is called \be coherent. \ee
Otherwise, $S$ does not
induce a set of handle assignments, and
is called \be incoherent. \ee
\end{definition}

The above definition does not guarantee that the assignments induced
by a coherent support set are admissible, that the same atom is not
required to be both true and false in the assignments of different
cycles, and that the incoming handles of a cycle being supposed to
be active/not active corresponds to a suitable setting of the
out-handles of the cycles they come from. Consider for instance
cycle $C_i$ which has an incoming handle, e.g. $h =(\beta : OR
:\lambda)$, in $IN^A_{C_i}$: $h$ is supposed to be active, which in
turn means that $\beta$ must be concluded true elsewhere in the
program; then, for all cycles $C_j$ where $\beta$ is involved into,
we must have $\beta \in OUT_{C_j}^{+}$, in order to fulfill the
requirement. Of course, we have to consider both $IN^A_C$ and
$IN_N^C$, and both the AND and the OR handles.

The following definition formalizes these more strict requirements.

\begin{definition}
\label{adequate} A coherent CG support set $S$ is \be adequate \ee
(w.r.t. not adequate) if for the induced handle assignments the
following conditions hold:
\begin{enumerate}
\item
they are admissible;
\item
for each two cycles
$C_i$, $C_j$ in $\Pi$,
$OUT_{C_i}^{+} \cap OUT_{C_j}^{-} = \emptyset$.
\item
For every $C_i$ in $\Pi$, for every handle $h \in IN^A_{C_k}$
of the form either $(\beta : OR : \lambda)$ or $(\no \beta : AND : \lambda)$,
and for every handle $h \in IN^N_{C_k}$
of the form either $(\beta : AND : \lambda)$ or $(\no \beta : OR : \lambda)$,
for every other cycle $C_j$ in $\Pi$, $i \neq j$,
such that $\beta \in Out\_handles(C_j)$, we have $\beta \in OUT_{C_j}^{+}$.
\item
For every $C_i$ in $\Pi$, for every handle $h \in IN^A_{C_k}$
of the form either $(\no \beta : OR : \lambda)$ or $(\beta : AND : \lambda)$,
and for every handle $h \in IN^N_{C_k}$
of the form either $(\no \beta : AND : \lambda)$ or $(\beta : OR : \lambda)$,
for every other cycle $C_j$ in $\Pi$, $i \neq j$,
such that $\beta \in Out\_handles(C_j)$, we have $\beta \in OUT_{C_j}^{-}$.
\end{enumerate}
\end{definition}

\section{Checking consistency on the Cycle Graph: Main Result}

We are now able to state the main result of the paper, which gives
us a necessary and sufficient syntactic condition for consistency
based on the Cycle Graph of the program.

\begin{theorem}
\label{mainresult}
A program $\Pi$ has stable models if and only if
there exists and adequate CG support set $S$ for $\Pi$.
\end{theorem}
\begin{proof}
\noindent
$\Leftarrow$\\
On the basis of  $S$ we can obtain the
corresponding induced handle assignments, that will be admissible
by the hypothesis that the $S$ is adequate.

From $S$ we can obtain a \be global handle assignment \ee $HA = \langle T_{HA};F_{HA} \rangle$
as follows.

\ni
\(T_{HA} = \{\alpha | \\
\ \ \ \ (\alpha : OR) \in ACT^{+}(S) \vee (\no \alpha : AND) \in
ACT^{+}(S) \vee\\
\ \ \ \ (\no \alpha : OR) \in ACT^{-}(S) \vee (\alpha : AND) \in
ACT^{-}(S)\}\)

\ni \(F_{HA} = \{\alpha |\\
\ \ \ \ (\alpha : AND) \in ACT^{+}(S) \vee (\no \alpha : OR) \in
ACT^{+}(S) \vee \\
\ \ \ \ (\alpha : OR) \in ACT^{-}(S) \vee (\no \alpha : AND) \in
ACT^{-}(S)\}\)

Since the $S$ is adequate, then by Definition\rif{adequate} we have
that (i)
 for each cycle $C_i$ in $\Pi$, $S$ induces an admissible handle
 assignment,
that (ii) $HA$ is consistent (i.e. $T_{HA} \cap F_{HA} =
\emptyset$), that (iii) $\forall \alpha \in T_{HA}$, $\alpha$ is
concluded true in every cycle $C_i$ it is involved into since
$\alpha \in OUT^{+}_{C_i}$, and the handle assignment induced by $S$
to $C_i$ is admissible, and that (iv) $\forall \alpha \in F_{HA}$,
$\alpha$ is concluded false in every cycle $C_j$ it is involved
into, since $\alpha \in OUT^{+}_{C_j}$, and the handle assignment
induced by $S$ to $C_j$ is admissible.

On the basis of $HA$, for each cycle $C_i$ in $\Pi$ we can build a
correspondent independent program $C_i + Aux(C_i) + X_i$, where we
let $X_i = Atoms(H_{C_i}) \cap T_{HA}$, and $X_i^{-} =
Atoms(H_{C_i}) \cap F_{HA}$. This independent program has a stable
model $S_i$ by construction, since the handle assignment induced by
$S$ to $C_i$ is admissible (point (i) above). This stable model is
(by Definition\rif{partialstable}) a partial stable model for $\Pi$
relative to $C_i$, with base $\langle X_i, X_i^{-} \rangle$. Taken
one $S_i$ for each $C_i$ in $\Pi$, in the terminology of
Definition\rif{compatible} the $S_i$'s constitute a compatible set
of partial stable models because, according to
Definition\rif{compatible}: (1) for any other cycle $C_j$, $X_i \cap
X_j^{-} = \emptyset$, since $T_{HA} \cap F_{HA} = \emptyset$ by
point (ii) above; (2) $\forall A \in X_i$, $A$ is concluded true in
some other cycle, by point (iii) above; (3) $\forall A \in X_i^{-}$,
$A$ is concluded false all the other cycles, by point (iv) above.
Then, by Theorem\rif{spartial}, $\Pi$ has stable models.

\noindent
$\Rightarrow$\\
If $\Pi$ has a stable model $M$, then by Theorem\rif{spartial} we
can decompose $M$ into a compatible set of partial stable models,
one partial stable model $S_i$, with base $\langle X_i, X_i^{-}
\rangle$, for each $C_i$ in $\Pi$. Since $X_i, X_i^{-} \subseteq
Atoms(H_{C_i})$, they correspond to sets $IN^A_{C_i}$ and
$IN^N_{C_i}$ of handles of $C_i$ that are made active/not active by
this base. By point (1) of Definition\rif{compatible}, for every two
cycles $C_i$, $C_j$ $X_i \cap X_i^{-} = \emptyset$, and then
$IN^A_{C_i} \cap IN^N_{C_j} = \emptyset$. If we let $S$ such that
$ACT^{+}(S) = \bigcup_{i \leq w} IN^A_{C_i}$ and $ACT^{-}(S) =
\bigcup_{i \leq w} IN^N_{C_i}$, we have $ACT^{+}(S) \cap ACT^{-}(S)
= \emptyset$ and, if there are either opposite or contrary handles,
they will not be in the same set. Then, $S$ is a $CG$ support set,
and is potentially adequate by construction, because it has been
built from the $IN^A$'s and $IN^N$'s of the cycles (point 1-2 of
Definition\rif{potadequate}), and because the $S_i$'s agree on
shared atoms, having been obtained by decomposing a stable model
(points 3-5 of Definition\rif{potadequate}). For the same reasons,
$S$ is also adequate.

\end{proof}

Checking the condition stated in Theorem\rif{mainresult} does not
imply finding the stable models of the program. However, in the
proof of the \textit{only-if} part, a way of determining the stable
models can be actually outlined, and is summarized below.

\begin{corollary}
\label{findstable}
Assume that the condition stated in Theorem\rif{mainresult} holds for program $\Pi$.
Then, the stable models of $\Pi$ can be determined as follows.
\begin{enumerate}
\item
Given the handle assignments induced by $S$,
build $T_{HA}$.
\item
For each cycle $C_i$ in $\Pi$, let $X_i = Atoms(H_{C_i}) \cap T_{HA}$,
build the corresponding extended cycle $C_i + Aux(C_i) + X_i$,
and find its partial stable models.
\item
Assemble each stable model of $\Pi$ as the union
of one partial stable model
for each cycle.
\end{enumerate}
\end{corollary}

Consider the following collection of cycles.

\st $
\begin{array}{l}
\li OC_0\\
p \ar \no s, \no c\\
s \ar \no t\\
t \ar \no p
\end{array} $

\st $
\begin{array}{l}
\li OC_1\\
p \ar \no s, \no c\\
s \ar \no t\\
t \ar \no p\\
s \ar a
\end{array} $

\st $
\begin{array}{l}
\li OC_2\\
q \ar \no q\\
q \ar \no e\\
\end{array} $

\st $
\begin{array}{l}
\li OC_3\\
r \ar \no r, \no e
\end{array} $

\st $
\begin{array}{l}
\li EC_1\\
a \ar \no c\\
c \ar \no a
\end{array} $

\st $
\begin{array}{l}
\li EC_2\\
a \ar \no b\\
b \ar \no a
\end{array} $

\st $
\begin{array}{l}
\li EC_3\\
e \ar \no f\\
f \ar \no e
\end{array} $

\st $
\begin{array}{l}
\li OC'_2\\
q \ar \no q\\
q \ar \no e\\
q \ar a
\end{array} $

Let $\pi_1 = OC_0 \cup EC_1$, $\pi_2 = OC_1 \cup EC_1 \cup EC_2$,
$\pi_3 = OC_1 \cup EC_1 \cup EC_2 \cup OC_2 \cup OC_3$, and $\pi_4 =
OC_1 \cup EC_1 \cup EC_2 \cup OC'_2 \cup OC_3$. The Cycle Graphs of
these programs are shown in Figure\rif{fig:CG1}, Figures 2, 3 and 4
respectively.

\begin{figure}[htbp]
    \begin{center}
    \epsfig{file=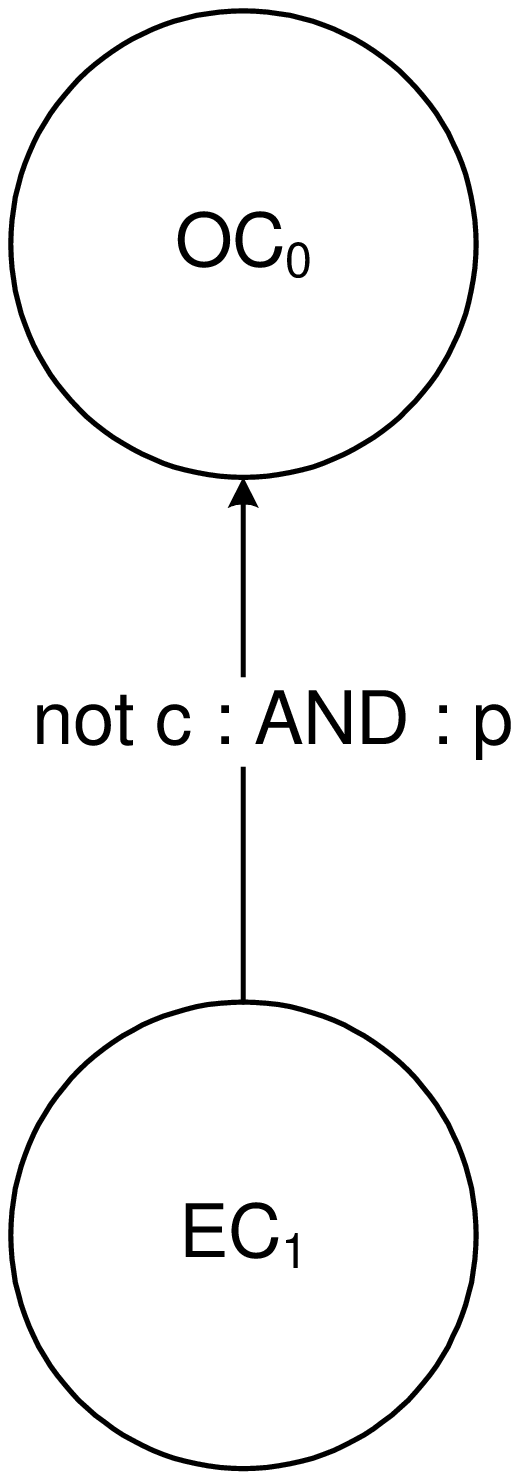, scale=0.4}
    \label{fig:CG1}
    \end{center}
    \caption{The Cycle Graph of $\pi_1$.}
\end{figure}

\begin{figure}[htbp]
    \begin{center}
    \epsfig{file=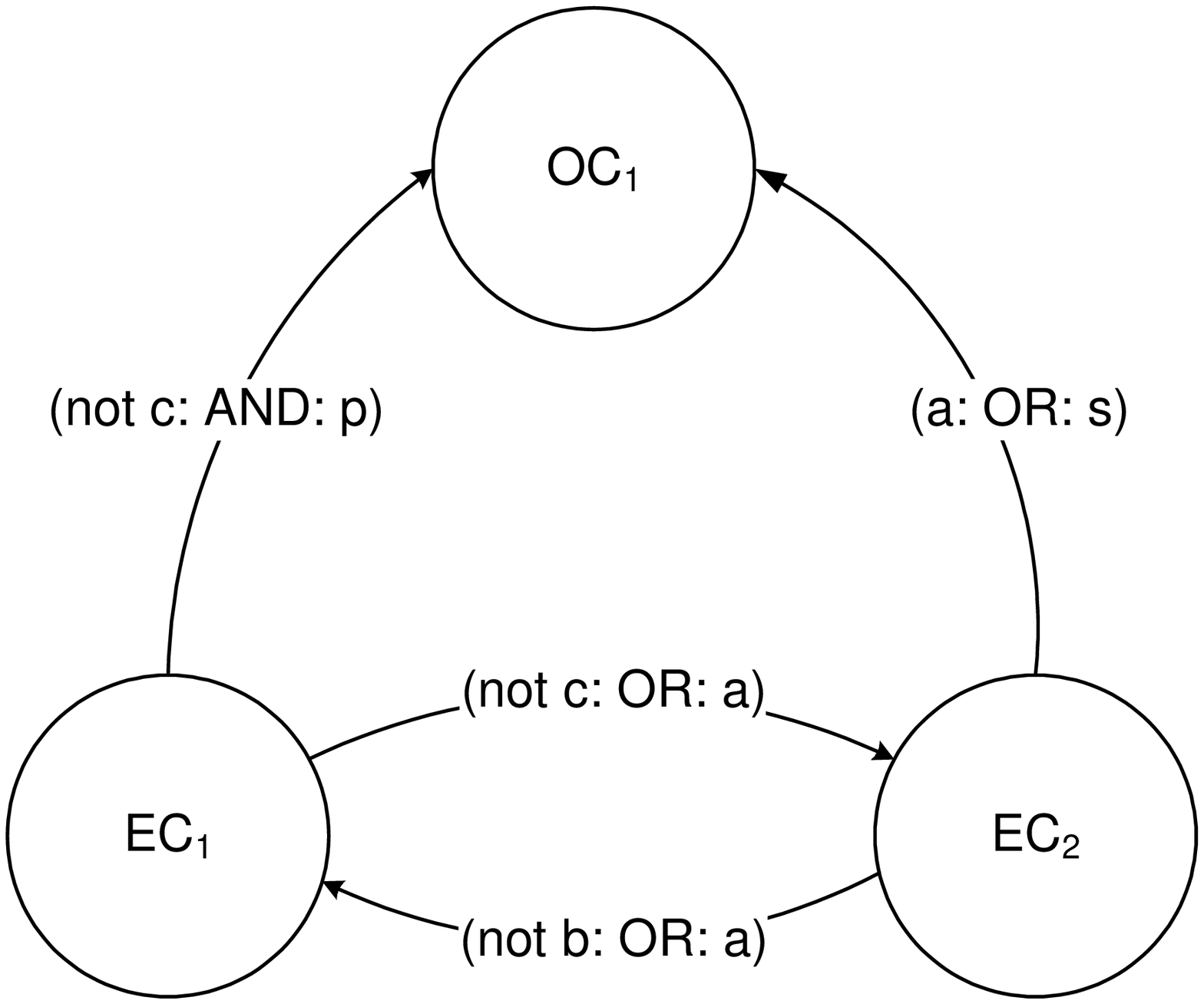, scale=0.4}
    \label{fig:CG2}
    \end{center}
    \caption{The Cycle Graph of $\pi_2$.}
\end{figure}

\begin{figure}[htbp]
    \begin{center}
    \epsfig{file=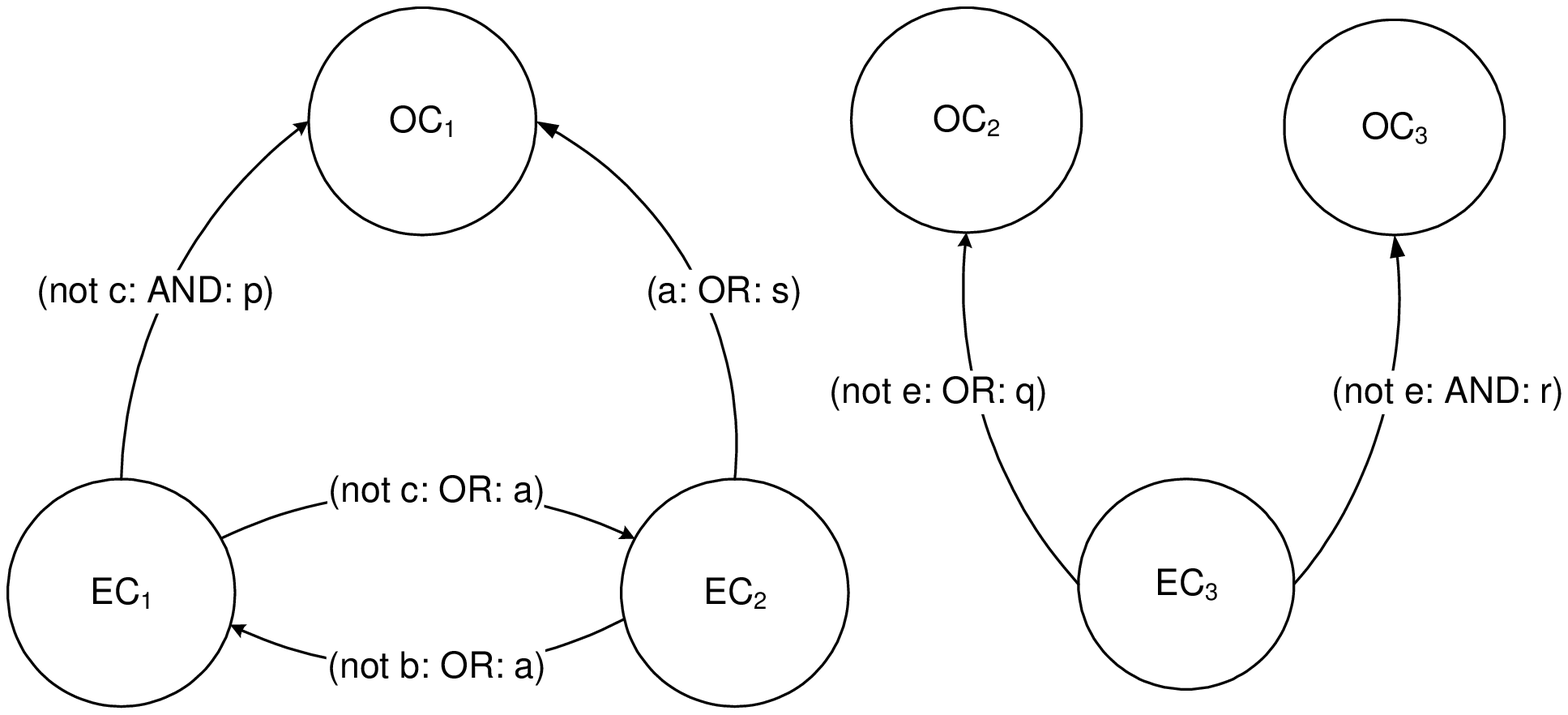, scale=0.7}
    \label{fig:CG3}
    \end{center}
    \caption{The Cycle Graph of $\pi_3$.}
\end{figure}

\begin{figure}[htbp]
    \begin{center}
    \epsfig{file=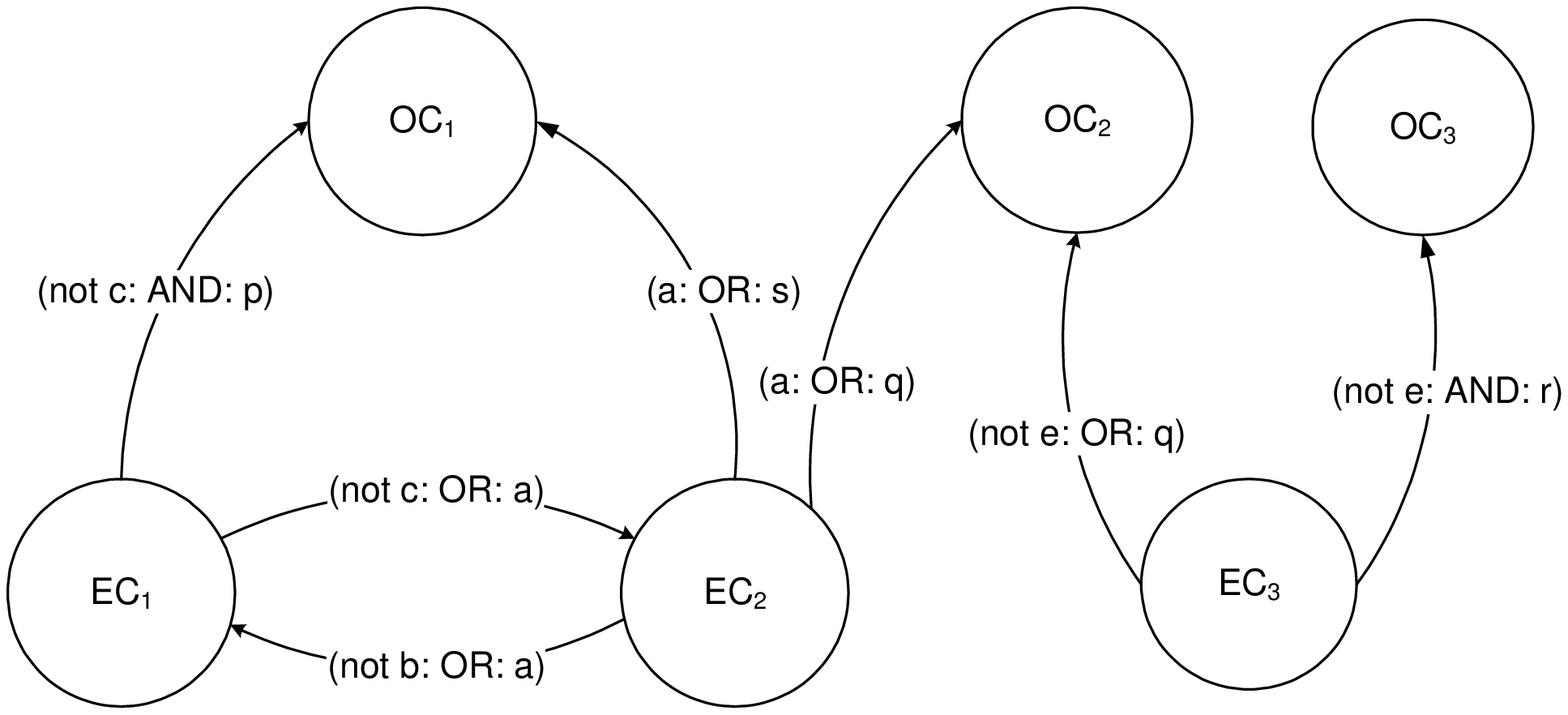, scale=0.7}
    \label{fig:CG4}
    \end{center}
    \caption{The Cycle Graph of $\pi_4$.}
\end{figure}

Let's denote $C + Aux(C)$ by $C^e$.

For $\pi_1$, we have $T_{HA} = \{c\}$. The partial stable model of
$EC_1^e \cup {c}$ is $\{c\}$; the partial stable model of $OC_0^e
\cup {c}$ is $\{c, t\}$. Then, a stable model of the overall program
is (as it is easy to verify) $\{c, t\}$.

For $\pi_2$, we have two possibilities.
In the first one, $T_{HA} = \{c\}$.
The partial stable model of $EC_1^e \cup {c}$
is $\{c\}$; the partial stable model of $EC_2^e \cup {c}$
is $\{c,b\}$; the partial stable model of $OC_1^e \cup {c}$ is
$\{c, t\}$. Then, a stable model of the overall program
is (as it is easy to verify) $\{c, b, t\}$.

In the second one, $T_{HA} = \{a\}$.
The partial stable model of $EC_1^e \cup {a}$
is $\{a\}$; the partial stable model of $EC_2^e \cup {a}$
is $\{a\}$; the partial stable model of $OC_1^e \cup {a}$ is
$\{a,s,t\}$. Then, a stable model of the overall program
is (as it is easy to verify) $\{a,s,t\}$.

For $\pi_3$ the situation is hopeless, since the only incoming
handles to $OC_2$ and $OC_3$ are opposite handles, that cannot be
both active.

For $\pi_4$, we have $T_{HA} = \{a,e\}$.
The partial stable model of $EC_1^e \cup {a}$
is $\{a\}$; the partial stable model of $EC_2^e \cup {a}$
is $\{a\}$; the partial stable model of $OC_1^e \cup {a}$ is
$\{a,s,t\}$; the partial stable model of ${OC'}_2^e \cup {a}$ is
$\{a,q\}$;  the partial stable model of $EC_3^e \cup {e}$ is
$\{e\}$; the partial stable model of $OC_3^e \cup {e}$ is
$\{e\}$.
Then, a stable model of the overall program
is (as it is easy to verify) $\{a,s,t,q,e\}$.

Therefore, it is useful to define a procedure for identifying
adequate support sets of a program on its Cycle Graph.

\section{Identifying adequate support sets on the Cycle Graph}
The above definitions allow us to define a procedure for trying to
\be find \ee adequate support sets starting from the odd cycles, and
following the dependencies on the $CG$.

\begin{definition}[Procedure PACG for finding adequate CG support sets for program $\Pi$]
\label{algo}
\begin{enumerate}
\item
Let initially $S = \langle \emptyset ; \emptyset \rangle$.
\item
For each cycle $C_k$ occurring in $\Pi$, $k \leq w$, let initially
$HA_{C_k} = \langle \emptyset, \emptyset, \emptyset, \emptyset
\rangle$
\item
For each odd cycle $OC$ in $\Pi$ do:
\begin{enumerate}
\item
Choose $h \in H_{OC}$. If $H_{OC} = \emptyset$, than FAIL.
\item For chosen $h$:
\begin{enumerate}
    \item do $ACT^{+}(S) := ACT^{+}(S) \cup \{h\}$;
    \item if $h^s$ occurs in the $CG$, do $ACT^{+}(S) := ACT^{+}(S) \cup \{h^s\}$;
    \item if $h^{-}$ occurs in the $CG$, do $ACT^{-}(S) := ACT^{-}(S) \cup \{h^{-}\}$;
    \item if $h^n$ occurs in the $CG$, do $ACT^{-}(S) := ACT^{-}(S) \cup \{h^n\}$.
\item
For each cycle $C_k$ in $\Pi$ such that $h \in H_{C_k}$:
\begin{enumerate}
    \item do $IN^A_{C_k} := IN^A_{C_k} \cup h$;
    \item if $h^s$ occurs in $H_{C_j}$ for some cycle $C_j$ (where possibly $j = k$), do
                $IN^A_{C_j} := IN^A_{C_j} \cup \{h^s\}$;
    \item if $h^{-}$ occurs in $H_{C_j}$ for some cycle $C_j$ (where possibly $j = k$), do
                $IN^N_{C_j} := IN^N_{C_j} \cup \{h^{-}\}$;
    \item if $h^n$ occurs in $H_{C_j}$ for some cycle $C_j$ (where possibly $j = k$), do
                $IN^N_{C_j} := IN^N_{C_j} \cup \{h^n\}$.
\end{enumerate}
\item
If $h$ is either of the form $(\beta : OR)$ or $(\no \beta : AND)$,
for each cycle $C_k$ in $\Pi$ where $\beta \in Out\_handles(C_k)$,
do: $OUT_{C_k}^{+} := OUT_{C_k}^{+} \cup \{\beta\}$;
\item
If $h$ is either of the form $(\no \beta : OR)$ or $(\beta : AND)$,
for each cycle $C_k$ in $\Pi$ where $\beta \in Out\_handles(C_k)$,
do: $OUT_{C_k}^{-} := OUT_{C_k}^{-} \cup \{\beta\}$
\end{enumerate}
\end{enumerate}
\item
\label{repeat} REPEAT
\begin{enumerate}
\item
Verify that $ACT^{+}(S) \cap ACT^{-}(S) = \emptyset$. If not, FAIL.
\item
Verify that neither $ ACT^{+}$ nor $ACT^{-}$ contain a pair of
either opposite or contrary handles. If not, FAIL.
\item
For each cycle $C_k$ in $\Pi$ such that $OUT_{C_k}^{+} \neq
\emptyset$ or $OUT_{C_k}^{-} \neq \emptyset$:
\begin{enumerate}
\item
Verify that $OUT_{C_k}^{+} \cap OUT_{C_k}^{-} = \emptyset$. If not,
FAIL.
\item
Update (if needed) $IN^A_{C_k}$ and $IN^N_{C_k}$ w.r.t.
$OUT_{C_k}^{+}$ and $OUT_{C_h}^{-}$, and check that the resulting
handle assignment is admissible. If not, then FAIL.
\item
For each other cycle $C_h$ in $\Pi$ do: verify that $OUT_{C_k}^{+}
\cap OUT_{C_h}^{-} = \emptyset$, and that $OUT_{C_k}^{-} \cap
OUT_{C_h}^{+} = \emptyset$. If not, FAIL.
\end{enumerate}
\item For each cycle $C_k$ in $\Pi$, for each $h \in IN^A_{C_k}$:
\begin{enumerate}
    \item do $ACT^{+}(S) := ACT^{+}(S) \cup \{h\}$;
    \item if $h^s$ occurs in the $CG$, do $ACT^{+}(S) := ACT^{+}(S) \cup \{h^s\}$;
    \item if $h^{-}$ occurs in the $CG$, do $ACT^{-}(S) := ACT^{-}(S) \cup \{h^{-}\}$;
    \item if $h^n$ occurs in the $CG$, do $ACT^{-}(S) := ACT^{-}(S) \cup \{h^n\}$.
\item
For each cycle $C_h$ in $\Pi$ such that $h \in H_{C_h}$:
\begin{enumerate}
    \item do $IN^A_{C_h} := IN^A_{C_h} \cup h$;
    \item if $h^s$ occurs in $H_{C_j}$ for some cycle $C_j$ (where possibly $j = h$), do
                $IN^A_{C_j} := IN^A_{C_j} \cup \{h^s\}$;
    \item if $h^{-}$ occurs in $H_{C_j}$ for some cycle $C_j$ (where possibly $j = h$), do
                $IN^N_{C_j} := IN^N_{C_j} \cup \{h^{-}\}$;
    \item if $h^n$ occurs in $H_{C_j}$ for some cycle $C_j$ (where possibly $j = h$), do
                $IN^N_{C_j} := IN^N_{C_j} \cup \{h^n\}$.
\end{enumerate}
\item
If $h$ is either of the form $(\beta : OR)$ or $(\no \beta : AND)$,
for each cycle $C_k$ in $\Pi$ where $\beta \in Out\_handles(C_k)$,
do: $OUT_{C_k}^{+} := OUT_{C_k}^{+} \cup \{\beta\}$;
\item

If $h$ is either of the form $(\no \beta : OR)$ or $(\beta : AND)$,
for each cycle $C_k$ in $\Pi$ where $\beta \in Out\_handles(C_k)$,
do: $OUT_{C_k}^{-} := OUT_{C_k}^{-} \cup \{\beta\}$
\end{enumerate}
\item For each cycle $C_k$ in $\Pi$, for each $h \in IN^N_{C_k}$:
\begin{enumerate}
    \item do $ACT^{-}(S) := ACT^{-}(S) \cup \{h\}$;
    \item if $h^s$ occurs in the $CG$, do $ACT^{-}(S) := ACT^{-}(S) \cup \{h^s\}$;
    \item if $h^{-}$ occurs in the $CG$, do $ACT^{+}(S) := ACT^{-}(S) \cup \{h^{-}\}$;
    \item if $h^n$ occurs in the $CG$, do $ACT^{+}(S) := ACT^{+}(S) \cup \{h^n\}$.
\item
For each cycle $C_h$ in $\Pi$ such that $h \in H_{C_h}$:
\begin{enumerate}
    \item do $IN^N_{C_h} := IN^N_{C_h} \cup h$;
    \item if $h^s$ occurs in $H_{C_j}$ for some cycle $C_j$ (where possibly $j = h$), do
                $IN^N_{C_j} := IN^N_{C_j} \cup \{h^s\}$;
    \item if $h^{-}$ occurs in $H_{C_j}$ for some cycle $C_j$ (where possibly $j = h$), do
                $IN^A_{C_j} := IN^A_{C_j} \cup \{h^{-}\}$;
    \item if $h^n$ occurs in $H_{C_j}$ for some cycle $C_j$ (where possibly $j = h$), do
                $IN^A_{C_j} := IN^A_{C_j} \cup \{h^n\}$.
\end{enumerate}
\item
If $h$ is either of the form $(\beta : OR)$ or $(\no \beta : AND)$,
for each cycle $C_k$ in $\Pi$ where $\beta \in Out\_handles(C_k)$,
do: $OUT_{C_k}^{-} := OUT_{C_k}^{-} \cup \{\beta\}$;
\item
If $h$ is either of the form $(\no \beta : OR)$ or $(\beta : AND)$,
for each cycle $C_k$ in $\Pi$ where $\beta \in Out\_handles(C_k)$,
do: $OUT_{C_k}^{+} := OUT_{C_k}^{+} \cup \{\beta\}$
\end{enumerate}
UNTIL no set is updated by the previous steps.
\end{enumerate}
\end{enumerate}
\end{definition}

\begin{proposition}
Procedure PACG either fails, or returns an adequate CG support set.
\label{correct_alg}
\end{proposition}

\begin{proof}
Whenever it does not fail, PACG clearly produces a CG support set
$S$. In fact: points (i-ii) of Definition\rif{CGSS} are verified by
steps 4.(a-b) of PACG; points (iii-v) of Definition\rif{CGSS} are
enforced after any update to $S$, namely by steps 3.b.(ii-iv),
4.d.(ii-iv) and 4.e.(ii-iv). The CG support set $S$ produced by PACG
is potentially adequate by construction, since in step 3.a a handle
for each odd cycle is included. $S$ is also adequate, since in fact:
admissible handle assignments for all cycles in $\Pi$ are
incrementally built and verified in steps 4.c.(i-ii), thus
fulfilling point 1. of Definition\rif{adequate}. Point 2 of
Definition\rif{adequate} is verified in step 4.c.(iii). Finally,
points 3-4 of Definition\rif{adequate} are enforced by steps
3.b.(vi-vii), 4.d.(vi-vii) and 4.e.(vi-vii), after each update to
the $IN_C$'s of any cycle.
\end{proof}

Let us reconsider collection of cycles given in previous sections,
and programs $\pi_1$, $\pi_2$, $\pi_3$ and $\pi_4$.

For $\pi_1 = OC_0 \cup EC_1$, the odd cycle $OC_0$ admits the unique
potentially active handle $(\no c : AND : p)$ Then, we let
$S_{\pi_1}$ be such that $ACT^{+}(S_{\pi_1}) = \{(\no c : AND)\}$
and $ACT^{-}(S_{\pi_1})= \emptyset$. The induced set of handle
assignments are as follows.

For $OC_0$: $IN^A_{OC_0} = \{(\no c : AND)\}$, $OUT^{+}_{OC_0}$ $=$
$OUT^{-}_{OC_0}$ $=$ $\emptyset$. This assignment is trivially
admissible, since there is no requirement on the out-handles.

For $EC_1$: $OUT^{+}_{EC_1}$ $=$ $\{c\}$, $OUT^{-}_{EC_1}$ $=$
$\{\emptyset\}$. $IN_{EC_1}$ $=$ $\emptyset$, since $EC_1$ is
unconstrained. It is easy to verify that this handle assignment is
admissible, by letting $\lambda_1 = a$ and $\lambda_2 = c$, where of
course for $c$ to be true $a$ must be false. This handle assignment
corresponds to selecting the partial stable model $\{c\}$ for
$EC_1$, while discarding the other partial stable model $\{a\}$.

Then, $S_{\pi_1}$  as defined above is an adequate CG support set.

Consider program $\pi_2 = OC_1 \cup EC_1 \cup EC_2$. The situation
here is complicated by the fact that $EC_1$ and $EC_2$ are not
independent. In fact, rule $a \ar \no b$ of $EC_2$ is an auxiliary
rule for $EC_1$, and, vice versa, $a \ar \no c$ of $EC_1$ is an
auxiliary rule for $EC_2$. Then, here we have a cyclic connection
between the even cycles. This is evident on the cycle graph of
$\pi_2$, reported in Figure 2.

The odd cycle $OC_1$ has two handles, of which at least one must be
active. Let us first assume that $(\no c : AND : p)$ is active.
According to the PACG procedure, we try to assemble a CG support set
$S$, by letting at first $ACT^{+}(S_{\pi_2}) = \{(\no c : AND)\}$
and $ACT^{-}(S_{\pi_2}) = \{(\no c : OR)\}$. In fact, since $\no c$
is an incoming OR handle for $a$ in $EC_2$, when assuming $(\no c :
AND)$ to be active, we also have to assume its opposite handle and
its contrary handle to be not active.

Accordingly, we let $IN^A_{OC_1} = \{(\no c : AND)\}$ and
$IN^N_{EC_2} = \{(\no c : OR)\}$ Now, we have to put $OUT^{-}_{OC_1}
= \{p\}$ and $OUT^{+}_{EC_1} = \{c\}$. To form an admissible handle
assignment for $EC_1$, this implies to let $IN^N_{EC_1} = \{(\no b :
OR)\}$. Consequently, we have to update $ACT^{-}(S_{\pi_2})$ which
becomes: $ACT^{-}(S_{\pi_2}) = \{(\no c : AND), (\no b : OR)\}$.
This leads to put $OUT^{+}_{EC_1} = \{b\}$.

Further iteration of the procedure changes nothing, and thus the
pair of sets $ACT^{+}(S_{\pi_2}) = \{(\no c : AND)\}$ and
$ACT^{-}(S_{\pi_2}) = \{(\no c : OR), (\no b : OR)\}$ form, as it is
easy to verify, an adequate CG support set.

Notice that this kind of reasoning requires neither to find the
stable models of the cycles, nor to consider every edge of the $CG$.
In fact, we do not need to consider the second incoming handle of
$OC_1$.

Let us now make the alternative assumption, i.e. assume that $(a :
OR : s)$ is active for $OC_1$. This means at first
$ACT^{+}(S_{\pi_2}) = \{(a : OR)\}$ and $ACT^{-}(S_{\pi_2}) =
\emptyset$, since $\no a$ does not occur in handles of the $CG$.
This implies $OUT^{+}_{EC_1} = \{a\}$. Thus, there is no requirement
on $IN_{EC_2}$ for forming an admissible handle assignment, and then
the procedure stops here.

For program $\pi_3 = OC_1 \cup EC_1 \cup EC_2 \cup OC_2 \cup OC_3$,
as we have already seen the only incoming handles to $OC_2$ and
$OC_3$ are opposite handles, that cannot be both active. For the
other cycles, the situation is exactly as before.

Then, there is a subprogram which gives problems. We can fix these
problems for instance by replacing $OC_2$ with $OC'_2$, thus
obtaining program $\pi_4$ (CG in Figure 4) where we can exploit
handle $(a : OR)$ for both $OC_1$ and $OC'_2$. It is easy to verify
that the CG support set $S$ composed of $ACT^{+}(S_{\pi_4}) = \{(a :
OR), (\no e : AND)\}$ and $ACT^{-}(S_{\pi_4}) = \{(\no e : OR)\}$ is
adequate. The need to support $OC'_2$ rules out the possibility of
supporting $OC_1$ by means of the handle $(\no c : AND : p)$.

\section{Usefulness of the results}

We believe that our results can be useful in different directions:
(i) Making consistency checking algorithms more efficient in the
average case. (ii) Defining useful classes of programs which are
consistent by construction. (iii) Checking properties of programs
statically and dynamically, i.e., when modifications and updates
affect the existence and the number of answer sets. (iv) Introducing
component-based software engineering principles and methodologies
for answer set programming: this by defining, over the $CG$, higher
level graphs where vertices are components, consisting of bunches of
cycles, and edges are the same of the $CG$, connecting components
instead of single cycles.

\subsection{Splitting consistency checking into stages}

The approach and the result that we have presented here can lead to
defining new algorithms for computing stable models. However,
they can also be useful for improving existing algorithms.

We have identified and discussed in depth two
aspects of consistency checking: (1) the odd cycles must be (either directly or
indirectly) supported by the even cycles; (2) this support must
be consistent, in the sense that no contrasting assumptions on the
handles can be made.

Point (1) is related to the \as coarse'' structure of the program,
and can be easily checked on the $CG$, so as to rule out a lot of
inconsistent programs, thus leaving only the \as potentially
consistent'' ones to be checked w.r.t. point (2). This is the aspect
that might be potentially exploited by any approach to stable models
computation.

Notice that a CG support set $S$ determines a subgraph of the CG,
which is composed of all the edges (and the corresponding end vertices)
marked with the handles which occur in $S$.

\begin{definition}
Given the $CG$ of program $\Pi$, and a CG support set $S$,
an \be adequate support subgraph \ee is
a subgraph $CG_S$ of the $CG$, composed of the edges marked by
the handles belonging to $ACT^{+}(S)$ and $ACT^{-}(S)$, and of
the vertices connected by these edges.
\end{definition}

It is easy to see that, syntactically, $CG_S$ is composed of a set
of \be handle paths,\,\ee that connect the odd cycles, through a
chain of handles, to the even cycles (or to cyclic bunches of even
cycles) that are able to support them. Each path may include more
than one odd cycle, while each odd cycle must occur in at least one
path.

Then, point 1 above may consist in checking whether a subgraph of the $CG$ with this
syntactic structure exists.
Point 2, however performed, in essence must check whether the handles marking the subgraph
constitute an adequate $CG$ support set.

Staying within the approach of this paper, one may observe that the
PACG procedure can easily be generalized for computing the stable
models by performing the two steps in parallel. In fact, PACG
actually tries to reconstruct the $CG_S$, starting from the odd
cycles and going backwards through the $CG$ edges to collect the
handles that form the set $S$. At each step however, the procedure
updates the handle assignments of the cycles and performs the
necessary checks to be sure to be assembling an adequate set $S$.
The extension would consist in computing the stable models of the
extended cycles instead of just the handle assignments, and perform
the computation on the whole $CG$.

\subsection{Defining classes of programs that are consistent by construction}

Based on the $CG$ it is possible to define syntactic restrictions
that, with a slight loss of expressivity, may ensure the existence
of stable models. Suitable restrictions might be enforced \as on
line'' by an automated tool, while the program is being written.
This can be made easier by limiting the number of handles each cycle
my have.

The definition of classes of programs suitable for ``interesting''
applications is a topic of further research, but it can be useful to
give some hints here.

In the literature, various sufficiency conditions have been defined (beyond
stratification) for existence of stable models.

\begin{itemize}
\item
Acyclic programs, by Apt and Bezem \cite{apt91};
\item
Signed programs, by Turner \cite{turner94};
\item
Call-consistent programs, order consistent programs, and negative
cycle free programs by Fages \cite{Fag94}.
\end{itemize}


For programs without classical negation, the classes of acyclic
programs, signed programs, negative cycle free and order consistent
programs are either included or coincide with the class of locally
stratified programs \cite{teodor:survey}, and have a unique stable
model that coincides with the well-founded model. In that case,
their canonical counterpart is the empty program. Call-consistent
programs do not contain odd cycles by definition.

We define below a new very simple class of programs that are
guaranteed to have stable models, broader than the above ones since
we admit odd cycles.

\begin{definition}
A program $\Pi$ is called {\em tightly-even-bounded} if it is either
call-consistent, or (if not) the following conditions hold: (i)
every odd cycle has just one handle; (ii) this handle comes from an
unconstrained even cycle; (ii) if there are two odd cycles whose
handles come from the same even cycle, then two handles that
originate in the same kind of node are of the same kind.
\end{definition}

The above condition is clearly very easily and directly checked on the $CG$,
and can be made clearly visible and understandable to a user, via
a graphical interface.
If you take any other existing graph representation, like e.g.
the $EDG$ \cite{BCDP99} \cite{Cos01}, that computes stable models as {\em graph colorings},
the check is of course possible, but is less easy and less direct.

It is easy to see that:

\begin{theorem}
Every tightly even-bounded program $P$ has stable models.
\end{theorem}
\begin{proof}
Even cycles the handles come from are unconstrained, and conflicting
handles are excluded by definition. Then, we can build an adequate
CG support set by just assuming the incoming handles of the odd cycles to be active.
\end{proof}

Simple as it is, this is a class wider than that of consistent
programs, which is easy to understand by programmers, and is
guaranteed to have stable models. Moreover, we can further enlarge
this class by allowing a tightly even-bounded \as core'' program to
have a \as top'', i.e. a set of definitions that do not contain
cycles, and possibly depend upon atoms of the \as core'' part. The
resulting class of program has a {\em generate} part, consisting of
the even cycles, a {\em test} part consisting of the odd cycles
which \as prune'' stable models, and a {\em conclude} part that
draws further consequences.

\subsection{Checking Properties of Programs}

By inspecting the structure of the $CG$ it is possible in principle
to detect whether a program is categorical, i.e. has a unique answer
set, and to estimate the number of the answer sets. In past work, we
have investigated the effects on the existence and the number of
answer sets after modifications to the program. In particular, after
asserting lemmas \cite{cos96}, and if adding new rules to the
program in the program development stage \cite{cip03}. The above
results can be reformulated, made uniform and extended by employing
the $CG$ for representing the program. Other static and dynamic
program properties may be investigated.

\subsection{Generalizing the $CG$ to components/agents}

A relevant topic is, in our opinion, that of defining software
engineering principles for Answer Set Programming.

Here we propose to define a program development methodology for
Answer Set Programming by defining, over the $CG$, higher level
graphs where vertices are \be components, \ee and edges are that of
the $CG$, but connect components instead of single cycles. We give
below a first informal description of what kind of methodology we
actually mean.

Let a \be component \ee $\cal C$ be a bunch of cycles.
It can be developed on its own, or it can be identified on the $CG$ of a larger program.
Similarly to a cycle however, $\cal C$ is not meant to be an independent
program, but rather it has incoming handles.

As we have seen, partial stable models of cycles
are characterized by handle assignments.
Analogously, a component will be characterized by a \be component interface \ee
\[IN^A_{\cal C},IN^N_{\cal C},OUT_{\cal C}^{+},OUT_{\cal C}^{-}\]
that is meant to be a specification of which values the incoming handles
may take, either in order to keep the component consistent, or in order to select
some of its stable models. The out-handles provide the other components
with a mean of establishing a connection with this one, i.e., they are
true/false atoms that can make the incoming handles of other components
active/not active as required.

Differently from cycles, components in general will not export \be
all \ee their active handles, but only those they want to make
visible and available outside.

Based on the interface, it is possible to connect components, thus building a
\be Component Graph \ee $Comp\_G$. On this new graph $Comp\_CG$, one can either add new consistent
components, or modify existing ones, and can check over the handle
paths if there are problems for consistency, and how to fix
them.

Referring to the previous example, in $\pi_3$ we have the component
$OC_1 \cup EC_1 \cup EC_2$ which is consistent, and the component
$OC_2 \cup OC_3 \cup EC_3$ which is instead
inconsistent. Then, we have a $Comp\_CG$ with two unconnected
vertices. We have shown how to fix the problem by adding a handle
to $OC_2$, i.e., by suitably connecting the two components on
the $Comp\_CG$.

In this framework, components may even be understood as independent agents,
and making a handle active to a component may be understood as sending a message to
the component itself. Consider the following example, representing a fragment of the
code of a \textit{controller} component/agent:

\st $
\begin{array}{l}
circuit\_ok \ar \no fault\\
fault \ar \no fault, \no test\_ok
\end{array} $

\ni where $test\_ok$ is an incoming handle, coming from a \textit{tester} component/agent.
As soon as the \textit{tester} will achieve $test\_ok$, this incoming handle will become
active, thus making the \textit{controller} consistent, and able to conclude $circuit\_ok$.

A formal definition of the methodology that we have outlined, and a
detailed study of the applications, are important future directions
of this research.

\end{sloppypar}


\end{document}